# An Efficient High-Dimensional Gene Selection Approach based on Binary Horse Herd Optimization Algorithm for Biological Data Classification

Niloufar Mehrabi[a*], Sayed Pedram Haeri Boroujeni[b], Elnaz Pashaei[c]

[a*, b] School of Computing, Clemson University, Clemson, SC, USA

Email: nmehrab@g.clemson.edu

Email: shaerib@g.clemson.edu

[c] Department of Software Engineering, Istanbul Aydin University, Istanbul, Turkey

Email: elnazpashaei@aydin.edu.tr

**Abstract:**

The Horse Herd Optimization Algorithm (HOA) is a new meta-heuristic algorithm based on the behaviors of horses at different ages. The HOA was introduced recently to solve complex and high-dimensional problems. This paper proposes a binary version of the Horse Herd Optimization Algorithm (BHOA) in order to solve discrete problems and select prominent feature subsets. Moreover, this study provides a novel hybrid feature selection framework based on the BHOA and a minimum Redundancy Maximum Relevance (MRMR) filter method. This hybrid feature selection, which is more computationally efficient, produces a beneficial subset of relevant and informative features. Since feature selection is a binary problem, we have applied a new Transfer Function (TF), called X-shape TF, which transforms continuous problems into binary search spaces. Furthermore, the Support Vector Machine (SVM) is utilized to examine the efficiency of the proposed method on ten microarray datasets, namely Lymphoma, Prostate, Brain-1, DLBCL, SRBCT, Leukemia, Ovarian, Colon, Lung, and MLL. In comparison to other state-of-the-art, such as the Gray Wolf (GW), Particle Swarm Optimization (PSO), and Genetic Algorithm (GA), the proposed hybrid method (MRMR-BHOA) demonstrates superior performance in terms of accuracy and minimum selected features. Also, experimental results prove that the X-Shaped BHOA approach outperforms others methods.

## 1. Introduction

In recent years, many researchers have used DNA microarray datasets to analyze thousands of genes simultaneously and correlate their expression with clinical phenotypes in cancer research [1, 2]. Since the microarray dataset contains numerous redundant genes and a limited number of instances, the feature selection technique could be crucial for choosing informative genes [3]. Feature Selection (FS) should be applied in machine learning as a pre-processing phase in order to get optimal output with short training times and low memory consumption [4]. FS plays a significant role in data mining [5] to solve various problems such as data classification[6], data clustering [7], image processing [8], text clustering [9], disaster management [10], and disease forecasting [11]. FS is generally classified into three major groups based on a variety of evaluation criteria, *i.e.*, filter method [12], wrapper model [13], and embedded technique [14].

Filter methods measure the relevance of features by their correlation with the dependent variable. Also, this technique uses statistical methods for the evaluation of a subset of features [15]. Filters select some features solely based on intrinsic characteristics of the training data, without involving any learning algorithm. As a result, they are significantly efficient, easily adaptable to more complex microarray datasets, and are readily applied using only a single application to produce results. Nevertheless, it has a tendency to choose subsets with a large number of features (even all of them), so a threshold is needed to select a subset. There are a variety of filter methods such as Relief [16], Information Gain [17], and Chi-square [18]. On the other hand, wrapper methods measure the usefulness of a subset of features by actually training a model on it [19]. However, wrapper methods are also computationally very costly. Wrapper algorithms can evaluate interactions between genes based on the classifier's accuracy. The wrapper methods achieve a higher rate of predictive accuracy than filter methods. By embedding methods within the machine learning algorithm, the process of feature selection is complete. In other words, they choose the features during model training, which is why they are referred to as embedded techniques [20].

The feature selection is defined as an NP-hard problem. For a given microarray dataset with N number of genes, $2^n$ possible subsets require to be evaluated to demonstrate the best gene subsets. Due to the search space expanding exponentially as the number of features increases, it is impossible to conduct an exhaustive search for the optimal feature subset in a high-dimensional space [21]. Since the classical feature selection methods have some limitations and drawbacks, meta-heuristic algorithms have been successfully used with some feature selection methods. Meta-heuristic algorithms are more computationally efficient in avoiding local minima and finding optimal features. Swarm Intelligence (SI) is a subset of meta-heuristic algorithms inspired by the behavior of birds, wolves, dragonflies, whales, and other animals. Improved Whale Optimization Algorithm (WOA) [22], Chimp Optimization Algorithm (ChOA) [23], Flying Squirrel Optimizer (FSO) [24], Coyote Optimization Algorithm (COA) [25], Harris Hawks Optimization (HHO) [26], Grasshopper Optimization Algorithm (GOA) [27], and Intelligent Dynamic Genetic Algorithm (IDGA) [28] are some recent works of SI that utilized for solving optimization problems.

Horse herd optimization Algorithm (HOA) is a new meta-heuristic approach that is based on swarm intelligence [29]. Generally, HOA mimics the behavior patterns of horse herds at different ages and applies to solving simple and complex optimization problems, especially high-dimensional examples. There are some general patterns of behaviors among horses that are including Grazing (G), Hierarchy (H), Sociability (S), Imitation (I), Defense mechanism (D), and Roam (R) [30–32]. The behavior patterns of horses can be used to balance the exploration and the exploitation phase. The suggested approach can provide the best possible response in the shortest time and with the least amount of computational cost.

In this paper, a binary version of HOA (BHOA) is proposed to solve discrete problems. The main idea is to map continuous search space into binary search space. To this aim, a novel X-shaped transfer function is considered [33]. The X-shaped transfer function has superior performance in comparison to popular S-shaped and V-shaped transfer functions. Furthermore, the MRMR method is utilized as a filter method in the initial step of the hybrid gene selection method (MRMR-BHOA) to reduce redundant and irrelevant genes due to gene expression datasets having a large number of genes [34]. The efficiency of BHOA is evaluated by two different classifiers called Naïve Bayes (NB) and Support Vector Machine (SVM) on five well-known datasets. Also, the 10-folds-Cross-Validation method is used to prove the effectiveness of each classifier. Eventually, SVM is selected as an evaluator of selected genes in the proposed hybrid approach because SVM performs better in high-dimensional spaces. In addition, SVM works well when the number of dimensions exceeds the number of samples. Finally, statistical analysis and convergence rate analysis are used to compare BHOA with some existing nature-inspired optimization methods such as PSO, GA, and GWO.

Therefore, the main contributions of this work can be summarized as follows:

- Suggesting binary version of Horse Herd Optimization Algorithm and applying it for solving gene selection problem.
- Examining the performance of nine transfer functions including X-Shaped, V-shaped, and S-Shaped on Horse Herd Optimization Algorithm to transfer continuous search space into the binary value.
- Proposing a hybrid filter/wrapper gene selection method based on the MRMR approach and binary Horse Herd Optimization Algorithm.

The remaining parts of this study are structured as follows: in section 2, we briefly review the previous works. Sections 3 and 4 elaborate the complete framework, containing the continuous HOA, the proposed binary version of HOA, the MRMR filter approach, the X-shaped transfer function, and the proposed hybrid approach (MRMR-BHOA) for feature selection. After that, the experimental setting, the achieved results, the comparisons against other approaches, and the discussions are reported in section 5. Finally, the conclusions are stated in the last section.

## 2. Literature review

Various meta-heuristic algorithms have been developed and used for feature selection, as presented in the literature. Generally, there are three types of meta-heuristic optimization algorithms: Evolutionary Algorithms (EA), physics-based algorithms, and Swarm Intelligence (SI) algorithms.

The first category, evolutionary algorithms, attempts to imitate biological natural laws, rules, and processes. Darwin's theories are the basis for these algorithms. A Genetic Algorithm (GA) is the first evolutionary algorithm [35]. Natural selection is simulated through genetic algorithms, so those species that can adapt to environmental change survive, reproduce and go on to generate more offspring. In simple words, the fittest individuals are chosen for reproduction using this algorithm in order to produce the next generation of offspring for solving problems. GA has capable of solving complex and non-linear problems, as well as multi-objective optimization issues. Furthermore, GA has some drawbacks like low performance and a tendency to stick to local minima. Some examples of EAs are Simulated Annealing (SA) [36], Genetic programming (GP) [37], Memetic Algorithm (MA) [38], and Gradient Evolution Algorithm (GEA) [39].

Another type of meta-heuristic algorithm, physics-based, is based on the physical laws, such as the laws of electromagnetic forces, inertia, magnetic fields, weight, etc [40]. Some of the most well-known algorithms are as follows: Vibrating Particles System (VPS) [41], Binary Multi-Verse Optimizer (MVO) [42], Binary Dragonfly Algorithm (BDA) [43], Binary Gravitational Search Algorithm (GSA) [44], and Ideal Gas Molecular Movement (IGMM) [45]. As an example, BMVO is a physics-based model that utilizes three concepts from cosmology: a wormhole, black hole, and white hole. These three concepts have been mathematically modeled for exploring, exploiting, and local search. Each solution in the BMVO is defined by binary values, with a V-Shaped function used to transform continuous values into binary values. The convergence speed of BMVO is extremely fast. Another popular physics-based algorithm is BGSA, which uses Newton's law of gravitation and mass interactions to find the optimal solution. Using stochastic rules, BGSA avoids local optimum and finds the most optimal solution [44]. As a result, the heaviest mass is chosen as the optimal solution of the search space because it attracts the remaining masses [46]. A binary GSA (BGSA) converts the result of gravitational forces into a probability number for each of the binary elements, which describes whether or not these elements will have the value 0 or 1.

The last subcategory of meta-heuristic optimization algorithms is Swarm Intelligence (SI). The SI algorithms imitate the behavior of animals, such as birds, bats, horses, wolves, or chimps in a group [47]. The most common swarm intelligence schemes are Binary Particle Swarm Optimization (BPSO) [48], Binary Bat Algorithm (BBA) [49], Binary Whale Optimization Algorithm (BWOA) [50], Flying Squirrel Optimizer (FSO) [24], Binary Harris Hawks Optimization (BHHO) [51], Binary Emperor Penguin Optimizer (BEPO) [52], Binary Dragonfly Algorithm (BDA) [43], Improved Binary Grey Wolf Optimization (IBGWO) [53], and Binary Coyote Optimization (COA) [20]. Eberhard and Kennedy [54] introduced particle swarm optimization (PSO) in 1995. The original PSO was based on biological social behavior, particularly the ability of some animal species to collaborate to locate food sources within an area. PSO is well-known for fast exploration and exploitation in the search space. The PSO algorithm starts with an initial random population of a candidate solution. Each individual of the population is called a particle [55]. The population moves around a multidimensional search space, with each particle representing a possible solution. Particles update their position based on their current velocity and the best position they have achieved so far. One of the benefits of PSO is its capability to provide high-quality solutions as quickly as possible and with a steady convergence characteristic [56].

Another SI algorithm is the Chimp Optimization Algorithm (ChOA) [57], which replicates the social dynamics of chimpanzees during group hunting, considering their individual intelligence and motivations. These groups consist of diverse chimpanzee types with distinct abilities and strategies for exploring their environment. The algorithm categorizes chimps into drivers, barriers, chasers, and attackers, each with unique skills for different stages of hunting. Drivers pursue prey, barriers obstruct movement, chasers follow and overtake, and attackers block escape routes. Attacker performance hinges on age, intelligence, and strength. Hunting involves exploration (driving, chasing, blocking) and exploitation (attacking, hunting), with ChOA's effectiveness contingent on striking the right balance between the two.

Emperor Penguin Optimizer (EPO) is another example of the meta-heuristic algorithms that mimic the emperor penguin's huddling behavior [58]. EPO consists of four primary steps: generating the huddle boundary, calculating the temperature around the huddle, computing the distance, and determining the most efficient mover. Considering that EPO cannot be used to solve discrete and binary problems, a binary version of EPO named BEPO was proposed [52]. In BEPO, the emperor penguin's location is transferred into binary search space using S-shaped and V-shaped transfer functions. The BEPO algorithm has also been used to solve the problem of feature selection.

Finally, the coyote optimization algorithm (COA), inspired by the Canis latrans species, has been proposed recently [25]. The COA takes into account the social organization of coyotes as well as their adaptation to their environment. It also introduces novel structures for balancing exploration and exploitation during the optimization process. The population size is defined by only two parameters: the number of packs and the count of coyotes within a pack. The COA method was proposed to solve global optimization problems with continuous values. Hence, a binary coyote optimization algorithm (BCOA) [20] was proposed to address feature selection problems in binary search space. The BCOA demonstrated excellent precision and a superb convergence curve.

## 3. Continuous Horse Herd Optimization Algorithm

The Horse Herd Optimization Algorithm was developed based on the behavior of horses in their natural environment at various ages [29]. In general, horse behavior can be divided into six categories: Grazing, Hierarchy, Sociability, Imitation, Defense Mechanism, and Roaming. Every horse moves according to the following equation during each iteration:

$$X_i^{\text{iter, AGE}} = V_i^{\text{iter, AGE}} + X_i^{\text{(iter-1),AGE}} \qquad , \qquad \text{AGE}=\alpha, \beta, \gamma, \delta \tag{1}$$

Where $X_i^{\text{iter, AGE}}$ demonstrates the location of the $i$th horse, $V_i^{\text{iter, AGE}}$ presents the velocity of the $i$th horse, *AGE* indicates the age range of each horse, and *iter* shows the current iteration. Horses' behavior changes at different ages. Each horse has a lifespan of 25–30 years on average. Regarding this, δ refers to the ages between 0 to 5 years, γ denotes the horses between the ages 5 and 10, β represents the age ranges from 10 to 15 years, and α shows the horses older than 15 years. The selection of horses' age should be based on a matrix of responses generated per iteration. Thus, the matrix can be arranged depending on which answers are the most effective, and according to Fig. 1 α horses are those horses ranked in the top 10% of the sorted matrix. The subsequent 20% belong to the β group. In terms of the remaining horses, the γ and δ horses consider 30% and 40%, respectively. The velocity of each horse can be calculated based on its age group and its behavior patterns within an iteration.

$$\begin{aligned}
V_i^{\text{iter},\alpha} &= G_i^{\text{iter},\alpha} + D_i^{\text{iter},\alpha} \\
V_i^{\text{iter},\beta} &= G_i^{\text{iter},\beta} + H_i^{\text{iter},\beta} + S_i^{\text{iter},\beta} + D_i^{\text{iter},\beta} \\
V_i^{\text{iter},\gamma} &= G_i^{\text{iter},\gamma} + H_i^{\text{iter},\gamma} + S_i^{\text{iter},\gamma} + I_i^{\text{iter},\gamma} + D_i^{\text{iter},\gamma} + R_i^{\text{iter},\gamma} \\
V_i^{\text{iter},\delta} &= G_i^{\text{iter},\delta} + I_i^{\text{iter},\delta} + R_i^{\text{iter},\delta}
\end{aligned} \tag{2}$$

In the following, we discuss how social and individual intelligence work for horses.

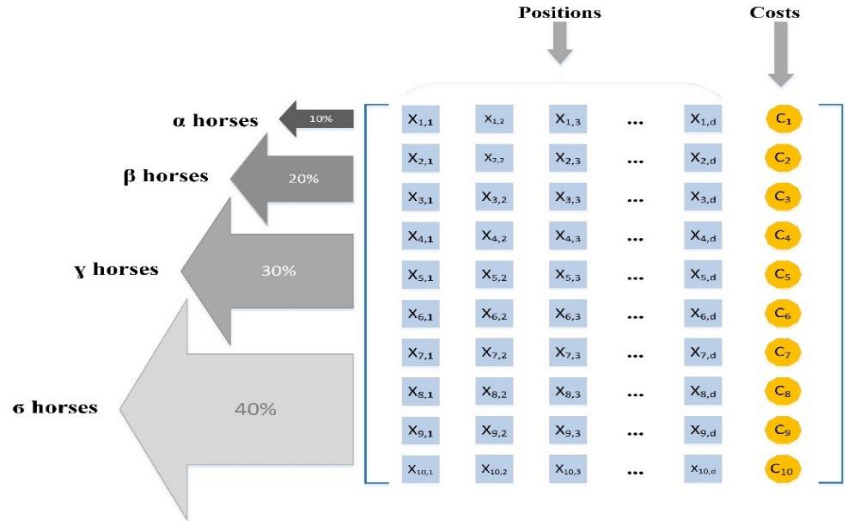

**Fig. 1** The Sorted Matrix of responses in the HOA algorithm

## A. Grazing (G)

Grazing is one of the most common horse behavior patterns. Horses graze roughly 70% of the time during the day and 50% of the time at night. In other words, at any age, they spend approximately 16 to 20 hours per day grazing on pastures. The grazing area around each horse is modeled by the HOA algorithm using the coefficient $g$ Equations (3) and (4) describe grazing mathematically.

$$G_i^{\text{iter, AGE}} = g_{\text{iter}}\left(\breve{u}+\mathcal{P}\breve{l}\right)\left[X_i^{(\text{iter-1})}\right] \qquad , \qquad AGE = \alpha, \beta, \gamma, \delta \qquad (3)$$

$$g_i^{\text{iter, AGE}} = g_i^{(\text{iter-1), AGE}}\times\omega_g \qquad (4)$$

Where $G_i^{\text{iter, AGE}}$ measures the tendency of the ith horse to graze that decreases linearly by $\omega_g$ with each iteration. Also, $\breve{l}$ and $\breve{u}$ indicate the lower and upper boundary of the grazing area, respectively, while $\mathcal{P}$ is a random value between 0 and 1. For all age groups, it is suggested [29] to assign 0.95 and 1.05 for $\breve{l}$ and $\breve{u}$, respectively, and the coefficient g is 1.5.

### B. Hierarchy (H)

There is always a hierarchy among horses because they are herd animals. The hierarchy protects horses while also allowing them access to better feeding grounds. In the HOA algorithm, the coefficient h represents the tendency of horses to follow the horse with the most experience and strength. Many investigations have shown that in the middle ages, β and γ, horses tend to have a hierarchical behavior. Equations (5) and (6) describe hierarchy behavior as follows:

$$H_i^{\text{iter, AGE}} = h_i^{\text{iter, AGE}}\left[X_*^{(\text{iter-1})}-X_i^{(\text{iter-1})}\right] \qquad , \qquad AGE = \alpha, \beta, \gamma \qquad (5)$$

$$h_i^{\text{iter, AGE}} = h_i^{(\text{iter-1), AGE}}\times\omega_h \qquad (6)$$

Where $X_*^{(\text{iter-1})}$ demonstrates the best horse's location, and $H_i^{\text{iter, AGE}}$ shows how the best horse's location affects the velocity parameter.

### C. Sociability (S)

Social behavior is a characteristic of horses. The purpose of sociability is to increase predator protection mechanisms and decrease scanning time. Furthermore, sociability increases intra-group competition and conflicts, disease transmission, and the risk of attracting predators. Considering flight is the top protection mechanism of horses, it is vital to identify hidden predators as soon as possible. Therefore, the ability to maintain shared awareness, act as a group during flight, and communicate effectively about such actions should be of prime importance for each horse of a herd. Observation shows that horses aged 5-15 are very interested in social life in a group. Factor s demonstrates the social behavior of horses, which is described by the following Equations (7) and (8).

$$S_i^{\text{iter, AGE}} = s_i^{\text{iter, AGE}} \left[ \left( \frac{1}{N} \sum_{j=1}^{N} X_j^{(\text{iter-1})} \right) - X_i^{(\text{iter-1})} \right] \quad , \quad \text{AGE} = \beta, \gamma \tag{7}$$

$$s_i^{\text{iter, AGE}} = s_i^{(\text{iter-1}), \text{AGE}} \times \omega_s \tag{8}$$

$S_i^{\text{iter, AGE}}$ denotes the social movement vector of the ith horse, and it decreases per iteration by the $\omega_s$ factor. Also, N is the total number of horses. AGE expresses the age group of horses. Furthermore, the average position is obtained as follows:

$$\text{Mean\_Position} = \left( \frac{1}{N} \sum_{j=1}^{N} X_j^{(\text{iter-1})} \right) \tag{9}$$

## D. Imitation (I)

Horses, as social animals, can learn from each other about good and bad behavior, such as discovering their suitable grassland [29]. In the age range of 0-5 years, young horses tend to imitate other horses. In addition, factor i in the HOA algorithm exhibits this feature of horses' behavior. Imitation in horse herds can be defined as follows:

$$I_i^{\text{iter, AGE}} = i_i^{\text{iter, AGE}} \left[ \left( \frac{1}{pN} \sum_{j=1}^{pN} \hat{X}_j^{(\text{iter-1})} \right) - X^{(\text{iter-1})} \right] \quad , \quad \text{AGE} = \gamma \tag{10}$$

$$i_i^{\text{iter, AGE}} = i_i^{(\text{iter-1}), \text{AGE}} \times \omega_i \tag{11}$$

$I_i^{\text{iter, AGE}}$ demonstrates the movement of the ith horse toward the mean position of the best horses, which are located at $\hat{X}$. pN indicates the number of horses in the best position. The recommended value of p is 10% of the total horses. Also, $\omega_i$ can be expressed as a reduction factor for each iteration, as mentioned previously. Moreover, the best location is obtained by the following equation.

$$\text{Best\_Position} = \left( \frac{1}{pN} \sum_{j=1}^{pN} \hat{X}_j^{(\text{iter-1})} \right) \tag{12}$$

## E. Defense Mechanism (D)

Horses have two primary defense mechanisms: flight and fight. When confronted with a dangerous situation, horses typically flee, and fighting serves as a secondary survival mechanism. Factor d describes the defense system of horses. Equations (13) and (14) with negative coefficients represent the horse's protective mechanisms, which prevent them from getting into inappropriate positions.

$$D_i^{\text{iter, AGE}} = - d_i^{\text{iter, AGE}} \left[ \left( \frac{1}{qN} \sum_{j=1}^{qN} \hat{X}_j^{(\text{iter}-1)} \right) - X^{(\text{iter}-1)} \right] \qquad , \qquad \text{AGE} = \alpha, \beta, \gamma \qquad (13)$$

$$d_i^{\text{iter, AGE}} = d_i^{(\text{iter}-1), \text{AGE}} \times \omega_d \qquad (14)$$

$D_i^{\text{iter, AGE}}$ denotes the flee vector of the ith horse based on the mean of horses with the worst areas. The number of horses with the worst locations is also shown by qN. It is proposed that q corresponds to 20 percent of all horses. As previously stated, $\omega_d$ represents the reduction factor for each iteration. Furthermore, the worst position is calculated as follows:

$$\text{Worst\_Position} = \left( \frac{1}{qN} \sum_{j=1}^{qN} \hat{X}_j^{(\text{iter}-1)} \right) \qquad (15)$$

### F. Roaming (R)

At the age range of 5-15 years, younger horses are more likely to move from pasture to pasture in looking for food. Horses are curious animals who seek out new pastures wherever they can. As horses reach maturity, they lessen their roaming behavior. The factor r is shown this behavior of horses as a random motion in Equations (16) and (17).

$$R_i^{\text{iter, AGE}} = r_i^{\text{iter, AGE}} \mathcal{P} X^{(\text{iter}-1)} \qquad , \qquad \text{AGE} = \gamma, \delta \qquad (16)$$

$$r_i^{\text{iter, AGE}} = r_i^{(\text{iter}-1), \text{AGE}} \times \omega_r \qquad (17)$$

In this case, ith horse's random velocity vector is represented by $R_i^{\text{iter, AGE}}$, and its reduction factor is indicated by $\omega_r$. To calculate the general velocity, the grazing, sociability, hierarchy, defense mechanism, imitation, and roaming are substituted into Equation (2).

By applying a sorting mechanism in a global matrix, HOA uses an appropriate technique for boosting the speed of issue resolution while also avoiding local optimal entrapment. The pseudo-code of HOA is stated in Algorithm 1. As indicated in Equations (18) and (19), the global matrix is constructed by juxtaposing positions (X) and the cost value for every position (C (X)).

$$X = \begin{bmatrix} x_{1,1} & x_{1,2} & x_{1,3} & \cdots & x_{1,d} \\ x_{2,1} & x_{2,2} & x_{2,3} & \cdots & x_{2,d} \\ \vdots & \vdots & \vdots & \ddots & \vdots \\ x_{m,1} & x_{m,2} & x_{m,3} & \cdots & x_{m,d} \end{bmatrix} \qquad , \qquad C\ (X) = \begin{bmatrix} c_1 \\ c_2 \\ \vdots \\ c_m \end{bmatrix} \qquad (18)$$

$$Global\ Matrix = [X\ C\ (X)] = \begin{bmatrix} x_{1,1} & x_{1,2} & x_{1,3} & \cdots & x_{1,d} & c_1 \\ x_{2,1} & x_{2,2} & x_{2,3} & \cdots & x_{2,d} & c_2 \\ \vdots & \vdots & \vdots & \ddots & \vdots & \vdots \\ x_{m,1} & x_{m,2} & x_{m,3} & \cdots & x_{m,d} & c_m \end{bmatrix} \tag{19}$$

---

**Algorithm 1**: Horse herd Optimization Algorithm

---

1:   Define input parameters
2:   Generate random positions for *n* horses
3:   Evaluate fitness of each horse's location
4:   Generate Global Matrix based on the horses' location and their fitness value
5:       **while** *(the stopping criterion is not satisfied)* **do**
6:           **for** i = 1: total number of horses **do**
7:               Sort the locations of horses in ascending order depending upon their fitness value
8:               Calculate the mean position by Equation (9)
9:               Calculate the good position by Equation (12)
10:              Calculate the bad position by Equation (15)
11:              Determining alpha, beta, gamma, and delta horses
12:              Computing the velocity of each horse by Equation (2)
13:              Update the position of each horse using Equation (1)
14:              Evaluate fitness for the new position of each horse
15:              **if** new fitness value < old fitness value **then**
16:                  set new position as the best position
17:                  set new fitness value as the best fitness value
18:              **end if**
19:          **end for**
20:      **end while**
21:  return the best position
22:  return the best fitness value

---

## 4. Proposed approach

### 4.1. Minimum Redundancy Maximum Relevancy (MRMR)

Hanchuan Peng has introduced the minimum redundancy maximum relevancy (MRMR) as a filter approach [59]. The MRMR attempts to determine the most significant attributes according to their correlation with the labels, as well as reducing the irrelevant features. This technique basically identifies the features that have maximum relevancy and minimum redundancy [34]. Mutual information (MI) quantifies both relevancy and redundancy by measuring the mutual dependence of two variables. A definition of mutual information is given by Equation (20).

$$\text{MI}(x_i, \text{C}) = \sum_{x \in X} \sum_{y \in Y} \text{P}(x_i, \text{C}) \log \frac{\text{P}(x_i, \text{C})}{\text{P}(x_i)\text{P}(\text{C})} \tag{20}$$

Where $\text{MI}(x_i, \text{C})$ shows the amount of mutual information between attribute x and the label of class c. $\text{P}(x_i)$ and $\text{P}(\text{C})$ present marginal probability functions, and $\text{P}(x_i, \text{C})$ indicates the joint probability distribution. If two random variables are completely independent, the mutual information value is 0. In the MRMR, the goal is to minimize redundancy (Rd) while maximizing relevance (Re). Using the maximum relevance, the top m features related to the class labels are selected.

$$\text{Re}(\text{S}) = \frac{1}{|\text{S}|} \sum_{x_i \in \text{S}} \text{MI}(x_i, \text{C}) \tag{21}$$

Where $\text{MI}(x_i, \text{C})$ is the mutual information of feature Xi with class C. A Minimum-Redundancy is used to eliminate redundancy between features which are described as follows:

$$\text{Rd}(\text{S}) = \frac{1}{|\text{S}|^2} \sum_{x_i, x_j \in \text{S}} \text{MI}(x_i, x_j) \tag{22}$$

Where $\text{MI}(x_i, x_j)$ is the mutual information of feature $x_i$ with $x_j$. A criterion that optimizes relevance and redundancy is called minimum redundancy maximum relevance (MRMR). The simplest way to optimize relevance and redundancy to gain an informative subset of features is as follows:

$$\max \emptyset \left( Re(\text{S}), Rd(\text{S}) \right) \tag{23}$$

where $\emptyset = (\text{Re}(\text{S}) - \text{Rd}(\text{S}))$.

### 4.2. The proposed Binary version of the Horse Herd Optimization Algorithm (BHOA)

In this paper, a binary version of HOA is introduced for feature selection problems, called BHOA. In general, discrete binary search spaces are necessary for various applications, such as feature selection. Additionally, problems with the continuous values could be turned into binary ones by transforming their variables to [0,1]. Any binary search space has a structure and limitations irrespective of the binary problem types. Each horse in the HOA moves in a continuous search space, with its position vector represented by continuous values. Accordingly, Equation (2) efficiently implements horse position updates through the addition of velocity variables to the position vector. In discrete problems, the solutions are limited to binary values like 0 and 1, so the positions cannot be updated using Equation (2). As a result, a method must be developed to change the horse's location by using velocity.

The proposed algorithm uses a Transfer Function (TF) as well as a method for updating the location. The TFs are divided into two groups based on their shapes: S-shaped and V-shaped. These two groups are illustrated in Fig. 2, and Table 1 shows the most commonly used mathematical formulations of the S-Shaped and V-Shaped TFs [60]. The transfer function produces a probability value according to the velocity of each horse, and this probability value allows continuous positions to convert into binary values.

**Table 1.** S-Shaped and V-Shaped transfer functions.

| S-Shaped | | V-Shaped | |
|---|---|---|---|
| Name | Transfer Function | Name | Transfer Function |
| $S_1$ | $T(v) = \dfrac{1}{1 + e^{-2x}}$ | $V_1$ | $T(v) = \left\| erf\left(\dfrac{\sqrt{\pi}}{2} v\right) \right\|$ |
| $S_2$ | $T(v) = \dfrac{1}{1 + e^{-x}}$ | $V_2$ | $T(v) = \|\tanh(v)\|$ |
| $S_3$ | $T(v) = \dfrac{1}{1 + e^{-(\frac{x}{2})}}$ | $V_3$ | $T(v) = \left\|(v)/\sqrt{1 + v^2}\right\|$ |
| $S_4$ | $T(v) = \dfrac{1}{1 + e^{-(\frac{x}{3})}}$ | $V_4$ | $T(v) = \left\|\dfrac{2}{\pi} arc\, tan\left(\dfrac{2}{\pi} v\right)\right\|$ |

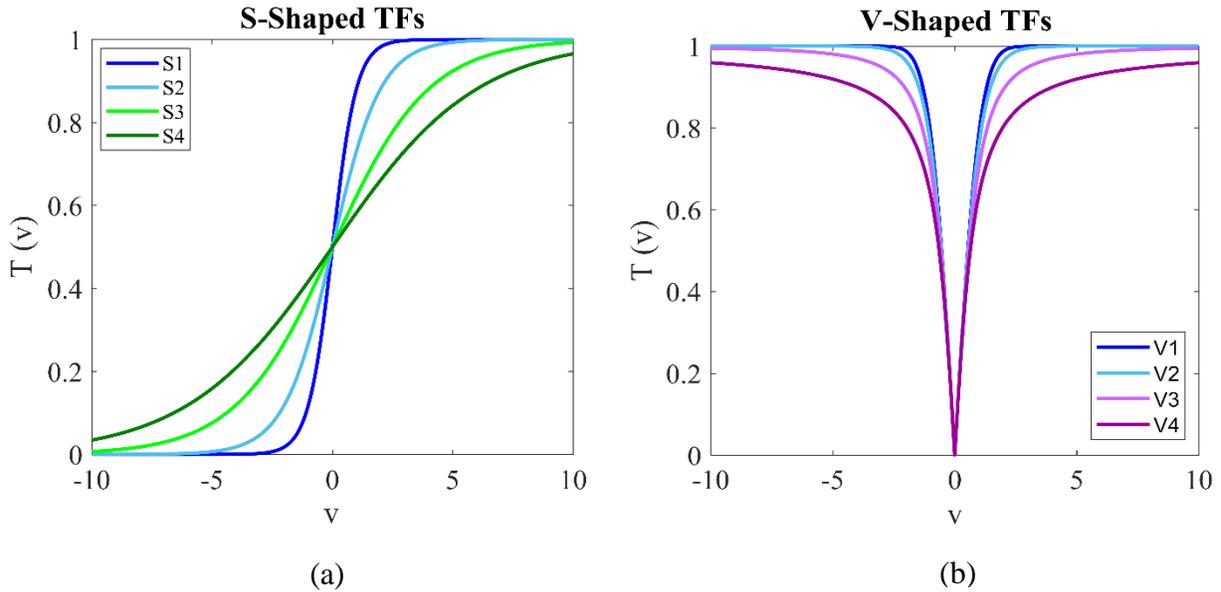

(a)                                                                 (b)

**Fig. 2** Transfer functions groups (a) S-Shaped and (b) V-Shaped

### 4.3. A novel X-Shaped Transfer Function

Due to the fact that the existing version of TFs (S-Shaped, V-Shaped) does not achieve an optimal balance between exploration and exploitation, we utilize a novel X-Shaped transfer function for updating the location of each horse [33]. As illustrated in Fig. 3, the X-Shaped TF includes two functions to enhance exploration and exploitation. Two new positions are generated using Equations (24) and (26). The best solution is selected by Equation (27), and it is then compared to the previous one.

$$W_1(v) = \frac{1}{1 + e^{-v}} \tag{24}$$

$$D_i(t+1) = \begin{cases} 1 & if\ rand_1 < W_1(t+1) \\ 0 & if\ rand_1 \geq W_1(t+1) \end{cases} \tag{25}$$

$$W_2(v) = \frac{1}{1 + e^{v}} \tag{26}$$

$$G_i(t+1) = \begin{cases} 1 & if\ rand_1 > W_2(t+1) \\ 0 & if\ rand_1 \leq W_2(t+1) \end{cases} \tag{27}$$

$$Z_i(t+1) = \begin{cases} D_i & if\ Fitness\ (D_i) < Fitness\ (G_i) \\ G_i & if\ Fitness\ (D_i) \geq Fitness\ (G_i) \end{cases} \tag{28}$$

Where $D_i$ and $G_i$ are the binary value obtained by Equation (24) and Equation (26) respectively, and $rand_1$ and $rand_2$ are random numbers between 0 and 1. If $Z_i(t+1)$ has a higher fitness value than the current position (X(i)), $Z_i$ will be considered as the new position. Otherwise, we employ a crossover operator on the current location (X(i)) and $Z_i$ to generate two children. The best crossover result determines the next position. The pseudo-code of X-shaped TF is shown in Algorithm 2.

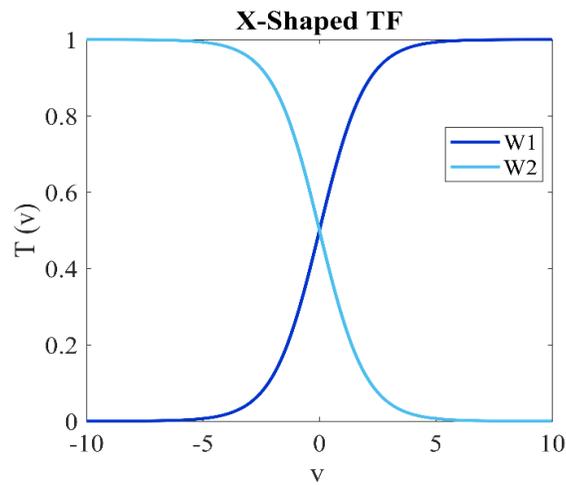

**Fig. 3** X-Shaped Transfer Function

---

**Algorithm 2**: X-Shaped Transfer Function

---

1:  **for** i = 1: total number of horses **do**
2:      Computing the velocity ($V_i$) of each horse by Equation (2)
3:      Calculate $W_1(V_i) = \frac{1}{1+e^{-v_i}}$
4:      **if** rand$_1$ < $W_1(V_i)$      **then**
5:              $D_i = 1$
6:      **else**
**7:**              $D_i = 0$
8:      **end if**
9:      Calculate $W_2(V_i) = \frac{1}{1+e^{v_i}}$
10:     **if** rand$_2$ < $W_2(V_i)$      **then**
11:             $G_i = 1$
12:     **else**
13:             $G_i = 0$
14:     **end if**
15:     **Calculate the new position X(t+1):**
16:     **if** f ($D_i$) < f ($G_i$)      **then**
17:             $Z$ (t+1) = $D_i$
18:     **else**
19:             $Z$ (t+1) = $G_i$
20:     **end if**
21:     **if** f (Z (t+1)) > f (X(t))      **then**
22:             $X$ (t+1) = $Z$ (t+1)
23:     **else**
24:             [child$_1$   ,   child$_2$] = Crossover (Z (t+1) ,  X(t))
25:             X(t+1) = best (child$_1$   ,   child$_2$)
26:     **end if**
27: **end for**

---

### 4.4. The proposed hybrid approach based on the BHOA for Gene Selection

In this section, we introduce MRMR-BHOA, a novel hybrid gene selection method that combines BHOA and MRMR to solve feature selection problems. The main aim of this work is to find an optimal gene subset regarding the maximum accuracy and the smallest number of genes. Furthermore, the proposed method significantly shortens time complexity and eliminates irrelevant genes. MRMR method is employed to minimize redundant genes in the initial stage, followed by BHOA which is applied to select the most efficient genes. Pseudo-code for the proposed algorithm can be found in algorithm 3.

Due to the problem involving whether to select or not select a given gene, if the horse position is zero, the corresponding gene is not chosen, and if the horse position is one, the corresponding gene is selected. The mechanism of selecting features is shown in Fig. 4. The SVM classifier evaluates the selected genes subset using the K-Fold-Cross Validation method. In BHOA, the fitness function is used to assess each horse's position. The fitness value is calculated according to the objective function below, considering the number of the selected genes and the classification accuracy. The objective function is described as follows:

$$\text{Fitness} = \alpha * ACC + \beta \frac{|N\text{-}S|}{|N|} \tag{29}$$

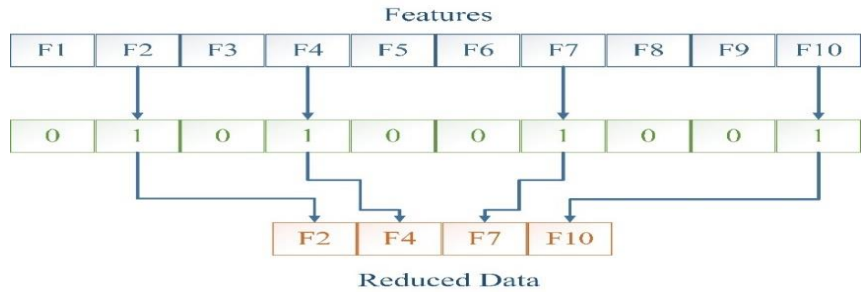

**Fig. 4** The mechanism of selecting features

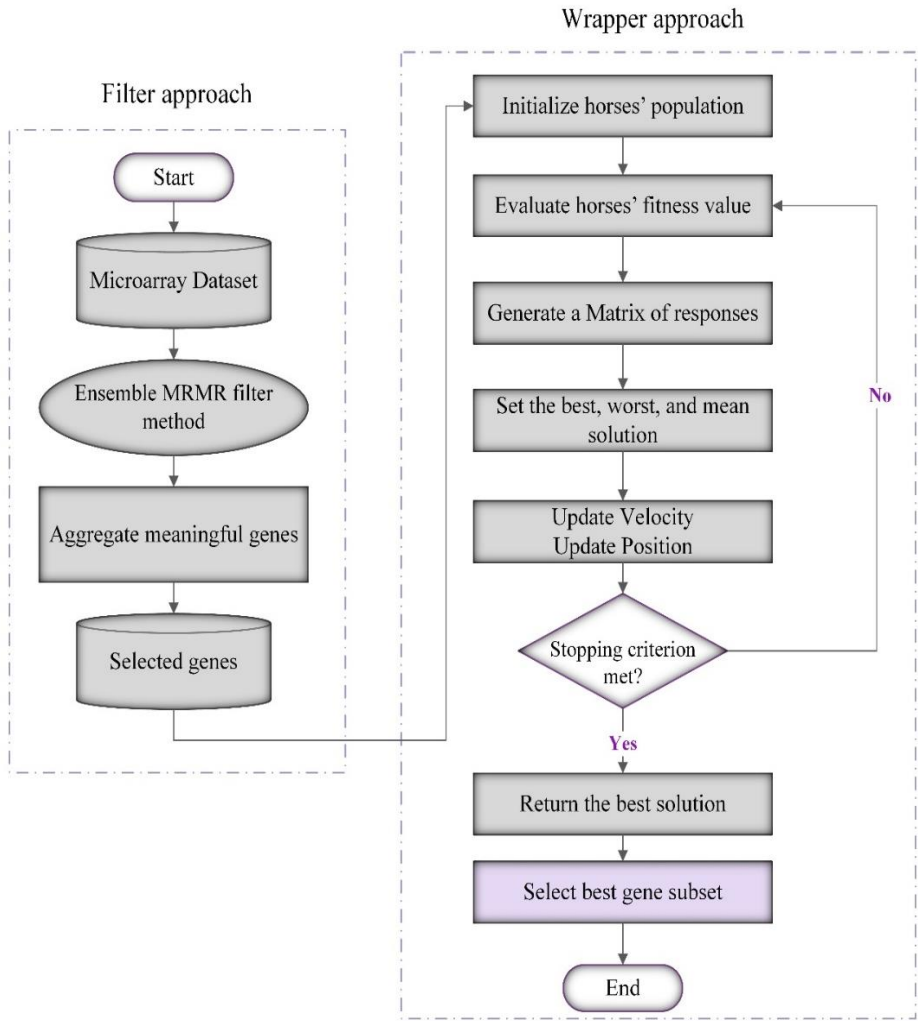

**Fig. 5** A framework of the proposed approach for the gene selection problem

Where, ACC indicates the accuracy of classification (obtained by using the SVM classifier), S and N are the number of selected genes and the total number of genes, respectively. Also, the two parameters, α, and β, correspond to the importance of classification quality and subset length, respectively. α is in the range [0, 1] and β = (1 − α). A framework of the proposed approach to the gene selection problem is illustrated in Fig. 5.

| **Algorithm 3**: MRMR-BHOA |
| --- |

1:    Define input parameters
2:    Generate random binary positions for *n* horses
3:    Evaluate fitness of each horse's location
4:    Generate Global Matrix based on the horses' location and their fitness value
5:    **while** *(the stopping criterion is not satisfied)*   **do**
6:        **for** i = 1: total number of horses   **do**
7:            Sort the locations of horses in ascending order depending upon their fitness value
8:            Calculate the mean position by Equation (9)
9:            Calculate the good position by Equation (12)
10:           Calculate the bad position by Equation (15)
11:           Determining alpha, beta, gamma, and delta horses
12:           Calculate   $W_1(V_i) = \frac{1}{1+e^{-v_i}}$
13:           **if** $rand_1 < W_1(V_i)$     **then**
14:               $D_i = 1$
15:           **else**
16:               $D_i = 0$
17:           **end if**
18:           Calculate   $W_2(V_i) = \frac{1}{1+e^{v_i}}$
19:           **if** $rand_2 < W_2(V_i)$     **then**
20:               $G_i = 1$
21:           **else**
22:               $G_i = 0$
23:           **end if**
24:           **Calculate the new position X(t+1):**
25:           **if** f ($D_i$) < f ($G_i$)     **then**
26:               Z (t+1) = $D_i$
27:           **else**
28:               Z (t+1) = $G_i$
29:           **end if**
30:           **if** f (Z (t+1)) > f (X(t))     **then**
31:               X (t+1) = Z (t+1)
32:           **else**
33:               [child$_1$  ,   child$_2$] = Crossover (Z (t+1) , X(t))
34:               X(t+1) = best (child$_1$  ,   child$_2$)
35:           **end if**
36:           **if** new fitness value < old fitness value     **then**
37:              set new position as the best position
38:              set new fitness value as the best fitness value
39:           **end if**
40:        **end for**
41:    **end while**
42:    return the best position
43:    return the best fitness value

## 5. Experimental result and discussion

### 5.1. Experimental setup

This section examined the proposed method on ten microarray datasets, consisting of DLBCL, Leukemia, SRBCT, Ovarian, Colon, Prostate, Lung, Brain-1, MLL, and Lymphoma that belong to the biomedical domain. Table 2 elaborates the details of each dataset concerning the number of genes and objects. The 10-Fold-Cross-Validation method, which randomly divides each dataset into two subsets called training and testing, was utilized to test the efficiency of the results. All approaches were repeated 20 times to achieve statistically significant results. Furthermore, the implementation of the proposed approaches was programmed using MATLAB 2017b, Intel Core i7, and 12 GB of RAM.

**Table 2.** Dataset description.

| Datasets | No. of genes | No. of Instances | No. of Classes |
|----------|--------------|------------------|----------------|
| Lymphoma | 4026 | 66 | 3 |
| Prostate | 10509 | 102 | 2 |
| Brain-1 | 5920 | 90 | 5 |
| DLBCL | 4026 | 47 | 2 |
| Colon | 2000 | 62 | 2 |
| Leukemia | 7129 | 72 | 2 |
| SRBCT | 2308 | 83 | 4 |
| Lung | 12600 | 203 | 5 |
| Ovarian | 15154 | 253 | 2 |
| MLL | 12582 | 72 | 3 |

### 5.2. Parameters tuning

The proposed MRMR-BHOA approach is compared to various widely used algorithms, such as Gray Wolf Optimization (GWO), Whale Optimization Algorithm (WOA), Ant Colony Optimization (ACO), Particle Swarm Optimization (PSO), Genetic Algorithm (GA), and Firefly algorithm. Listed below are the parameters set for competing algorithms in Table 3. To ensure fair comparisons, we assign a population size of 35 and a maximum number of iterations of 60, respectively. In addition, we chose $\alpha$ and $\beta$ in accordance with the literature (i.e., 0.99 and 0.01, respectively) [61]. Total features in a dataset determine the size of its search space. Furthermore, the efficiency of the results is also proven by statistical analysis.

**Table 3.** The parameters settings.

| Algorithm | Parameter | Explanation | Value |
|-----------|-----------|-------------|-------|
| BHOA | W | Reduction factor | 0.9 |
| | pN | Percent of best horses | 0.1 |
| | qN | Percent of worst horses | 0.2 |
| PSO | $C_1, C_2$ | Acceleration coefficient | 1.5, 2 |
| | W | Inertia weight | 0.9 |
| GA | $P_s$ | Selection mechanism | Roulette wheel |
| | $P_c$ | Crossover ratio | 0.7 |
| | $P_m$ | Mutation ratio | 0.1 |
| Firefly | α | Attractiveness coefficient | 0.2 |
| | β | Absorption coefficient | 1 |
| | ɣ | Light absorption coefficient | 1 |
| GWO | a | Coefficient | [2, 0] |
| ACO | α | Pheromone | 1 |
| | β | Visibility | 2 |
| | ρ | Evaporation of pheromone | 0.5 |
| WOA | a | Coefficient | [2, 0] |
| | b | Logarithmic spiral constant | 1 |

## 5.3. Measurement criteria

The classification accuracy is the first evaluation criterion used in this work. Generally speaking, the accuracy of classification measures indicates whether a classifier correctly identifies each features label. The results of these evaluations are computed based on a confusion matrix given in Table 4. Accuracy can be calculated by comparing rightly classified samples to the entire sample pool, as described below:

$$Accuracy = \frac{TP+TN}{TP+TN+FN+FP} \tag{30}$$

Where *TP* and *TN* represent true positives and true negatives, and false positives and false negatives are shown by *FP* and *FN*, respectively. As a result, we examine the capability of the proposed method using the SVM classifier which is used as an accuracy value in the fitness function. Matthews Correlation Coefficient (MCC), F-measure, and Area under ROC Curve (AUC) are also employed to assess the effectiveness of the proposed approach. These criteria are described by the following Equations:

$$MCC = \frac{TN \times TP \text{-} FN \times FP}{\sqrt{(TP+FP)+(TP+FN)+(TN+FP)+(TN+FN)}} \qquad (31)$$

$$F\text{-}Measure = 2 * \frac{Recall \times Precision}{Recall + Precision} \qquad (32)$$

**Table 4.** Confusion Matrix.

|  |  | Actual Condition | |
| --- | --- | --- | --- |
|  |  | Actual Positive | Actual Negative |
| Output of Classifier | Classify Positive | TP | FP |
|  | Classify Negative | FN | TN |

## 5.4. Experimental results and analysis

This paper compares the results of BHOA with recent approaches to assess the influence of the binary HOA and X-shaped transfer functions on feature selection. There are three main parts: Initially, we tested four different classifiers on datasets to determine which one would be the best to use as an evaluator for our approach. Then, employing and comparing various filter methods to select the most effective filter for dimensional reduction. The last step is to compare the X-Shaped TF to the S-Shaped and V-Shaped TF and demonstrate its significance. On ten biological datasets, four diverse classifiers (i.e., KNN, SVM, DT, and NB) were tested to determine which classification algorithm is more effective than the others. The comparison of various classifiers can be found in Table 5, and Figs. 6 and 7 illustrate the performance of different classifiers on ten datasets in terms of accuracy and F-measure. As a result of their close results, we decided to use SVM and KNN classifiers as evaluators during the filter method step.

### 5.4.1. Assessment of the performance of various filter techniques

In the following step, we compared the results of four different filter techniques to determine the most effective filter method for identifying informative genes to use as input to the proposed wrapper approach. As a result, we used the Relief, Chi-square, Laplacian, and MRMR methods to select the best 50 and 100 genes, which were then evaluated using the SVM and KNN classifiers. Table 6 presents a comparison of different filters. The finding results in Table 6 clearly show that the MRMR method outperforms other filter methods in terms of accuracy on all microarray datasets. Laplacian has the worst performance across all datasets. MRMR technique achieves higher classification accuracy on 7 out of 10 datasets for the top 50 genes. On the MLL, Leukemia, and Ovarian datasets, the best performance is obtained for the top 100 genes by the MRMR method. It should be noted that the SVM classifier with linear kernel achieves the highest efficiency on 8 out of 10 datasets, while the KNN classifier obtains better results on the remaining datasets (i.e. Colon, Prostate). Figs. 8 and 9 display a comparison of the performance of two different classifiers on the top 50 and 100 genes of each dataset by means of the MRMR filter method.

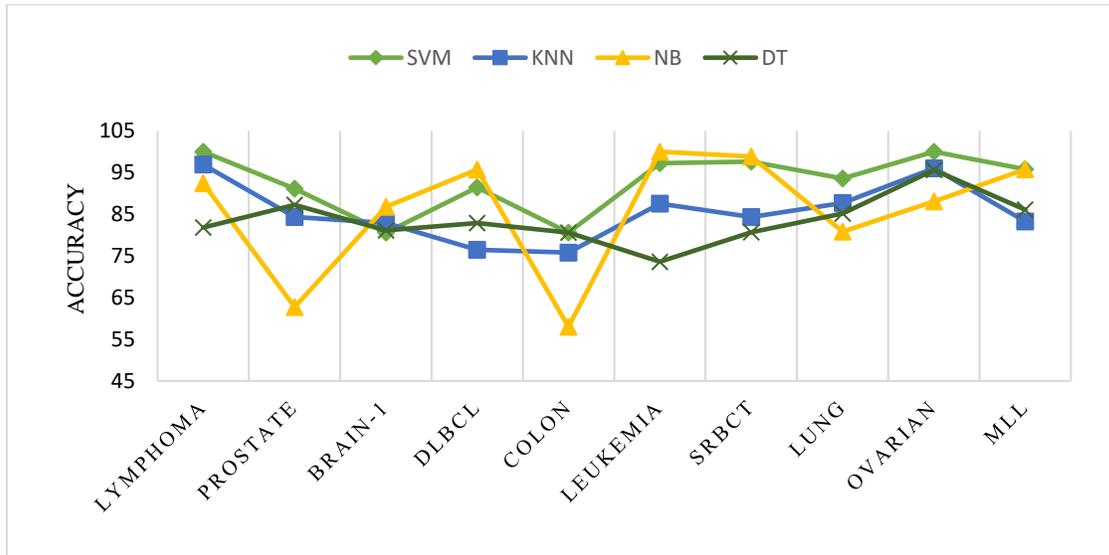

**Fig. 6** The performance of four classifiers on ten datasets in terms of accuracy

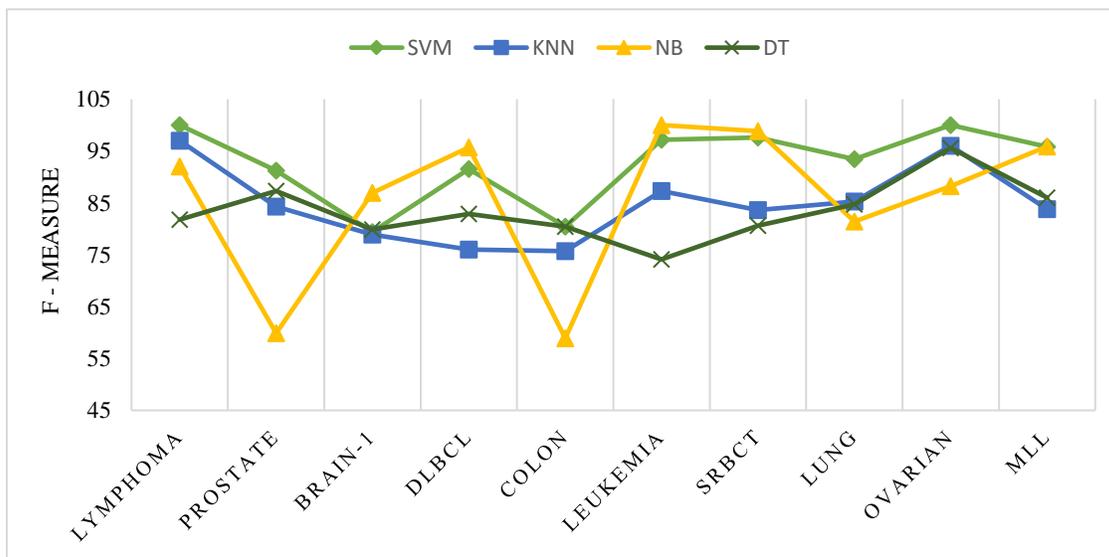

**Fig. 7** The performance of four classifiers on ten datasets in terms of F-measure

**Table 5.** The performance of four different classifiers on ten biological datasets.

| Dataset | Criteria | Classifier | | | |
|---------|----------|------|------|------|------|
| | | SVM | KNN | NB | DT |
| Lymphoma | Accuracy | 100 | 96.9 | 92.4 | 81.8 |
| | F-Measure | 100 | 97.0 | 92.0 | 81.8 |
| | MCC | 100 | 93.2 | 83.0 | 65.7 |
| | AUC | 100 | 95.4 | 87.5 | 89.8 |
| Prostate | Accuracy | 91.1 | 84.3 | 62.7 | 87.2 |
| | F-Measure | 91.2 | 84.3 | 59.9 | 87.3 |
| | MCC | 82.4 | 69.2 | 28.8 | 74.7 |
| | AUC | 91.2 | 84.4 | 62.7 | 97.6 |
| Brain-1 | Accuracy | 80.6 | 82.9 | 86.8 | 81.1 |
| | F-Measure | 79.4 | 78.9 | 86.9 | 79.9 |
| | MCC | 91.7 | 90.3 | 90.5 | 91.1 |
| | AUC | 79.4 | 78.8 | 86.8 | 79.8 |
| DLBCL | Accuracy | 91.4 | 76.5 | 95.7 | 82.9 |
| | F-Measure | 91.5 | 76.0 | 95.7 | 82.9 |
| | MCC | 83.3 | 55.4 | 91.5 | 66.1 |
| | AUC | 91.6 | 76.3 | 98.8 | 76.6 |
| Colon | Accuracy | 80.6 | 75.8 | 58.0 | 80.6 |
| | F-Measure | 80.4 | 75.7 | 58.9 | 80.4 |
| | MCC | 57.0 | 46.6 | 19.9 | 57.0 |
| | AUC | 77.8 | 73.1 | 64.8 | 66.5 |
| Leukemia | Accuracy | 97.2 | 87.5 | 100 | 73.6 |
| | F-Measure | 97.2 | 87.3 | 100 | 74.1 |
| | MCC | 93.9 | 72.0 | 100 | 44.9 |
| | AUC | 96.9 | 84.8 | 100 | 69.4 |
| SRBCT | Accuracy | 97.6 | 84.33 | 98.8 | 80.7 |
| | F-Measure | 97.6 | 83.6 | 98.8 | 80.6 |
| | MCC | 96.7 | 78.9 | 98.2 | 73.2 |
| | AUC | 98.1 | 96.9 | 99.1 | 87.5 |
| Lung | Accuracy | 93.6 | 87.7 | 80.8 | 85.2 |
| | F-Measure | 93.4 | 85.3 | 81.4 | 84.7 |
| | MCC | 86.1 | 94.8 | 65.5 | 71.1 |
| | AUC | 90.9 | 94.7 | 85.3 | 88.4 |
| Ovarian | Accuracy | 100 | 96.0 | 88.1 | 95.6 |
| | F-Measure | 100 | 96.0 | 88.2 | 95.6 |
| | MCC | 100 | 91.4 | 74.5 | 90.5 |
| | AUC | 100 | 95.5 | 93.0 | 95.7 |
| MLL | Accuracy | 95.8 | 83.3 | 95.8 | 86.1 |
| | F-Measure | 95.8 | 83.8 | 95.9 | 86.0 |
| | MCC | 93.8 | 76.7 | 93.7 | 78.8 |
| | AUC | 96.9 | 88.1 | 96.8 | 87.4 |

**Table 6.** Average classification accuracy (%) of top 50 and 100 genes obtained by different filter methods.

| Dataset | Classifier | 50 genes | | | | 100 genes | | | |
|---|---|---|---|---|---|---|---|---|---|
| | | MRMR | Relieff | Chi-square | Laplacian | MRMR | Relieff | Chi-square | Laplacian |
| Lymphoma | SVM | 100 | 100 | 98.48 | 96.90 | 100 | 100 | 98.48 | 98.33 |
| | KNN | 98.33 | 100 | 96.96 | 95.00 | 98.33 | 96.96 | 92.42 | 94.22 |
| Prostate | SVM | 92.24 | 92.16 | 90.19 | 65.57 | 91.84 | 95.09 | 88.23 | 68.46 |
| | KNN | 94.01 | 94.11 | 93.14 | 63.59 | 85.27 | 93.13 | 90.19 | 62.69 |
| Brain-1 | SVM | 84.44 | 83.26 | 84.73 | 68.18 | 90.52 | 88.72 | 90.12 | 75.73 |
| | KNN | 87.77 | 86.49 | 87.39 | 69.44 | 90.11 | 87.49 | 90.06 | 74.85 |
| DLBCL | SVM | 100 | 97.50 | 100 | 69.83 | 100 | 100 | 100 | 65.84 |
| | KNN | 95.74 | 100 | 95.74 | 61.33 | 100 | 100 | 95.74 | 60.83 |
| Colon | SVM | 83.87 | 83.87 | 83.91 | 73.90 | 82.14 | 85.36 | 79.23 | 72.62 |
| | KNN | 85.49 | 77.41 | 80.64 | 58.93 | 80.92 | 87.09 | 79.03 | 51.42 |
| Leukemia | SVM | 95.83 | 97.51 | 97.22 | 64.34 | 100 | 98.64 | 95.83 | 63.69 |
| | KNN | 93.05 | 96.07 | 93.25 | 73.15 | 98.75 | 98.61 | 97.22 | 73.15 |
| SRBCT | SVM | 100 | 97.63 | 98.79 | 81.12 | 100 | 100 | 100 | 89.16 |
| | KNN | 100 | 100 | 98.76 | 70.49 | 100 | 100 | 100 | 65.67 |
| Lung | SVM | 95.57 | 89.65 | 88.17 | 76.88 | 94.45 | 89.65 | 92.61 | 78.28 |
| | KNN | 95.56 | 92.61 | 93.10 | 68.47 | 94.16 | 93.59 | 93.10 | 68.05 |
| Ovarian | SVM | 99.61 | 98.83 | 98.81 | 90.87 | 99.62 | 98.02 | 100 | 90.41 |
| | KNN | 99.61 | 99.21 | 99.20 | 73.96 | 99.6 | 98.81 | 99.60 | 82.36 |
| MLL | SVM | 98.57 | 94.40 | 91.66 | 73.81 | 100 | 93.05 | 94.44 | 69.21 |
| | KNN | 98.46 | 98.46 | 95.83 | 58.51 | 100 | 95.83 | 93.05 | 58.75 |

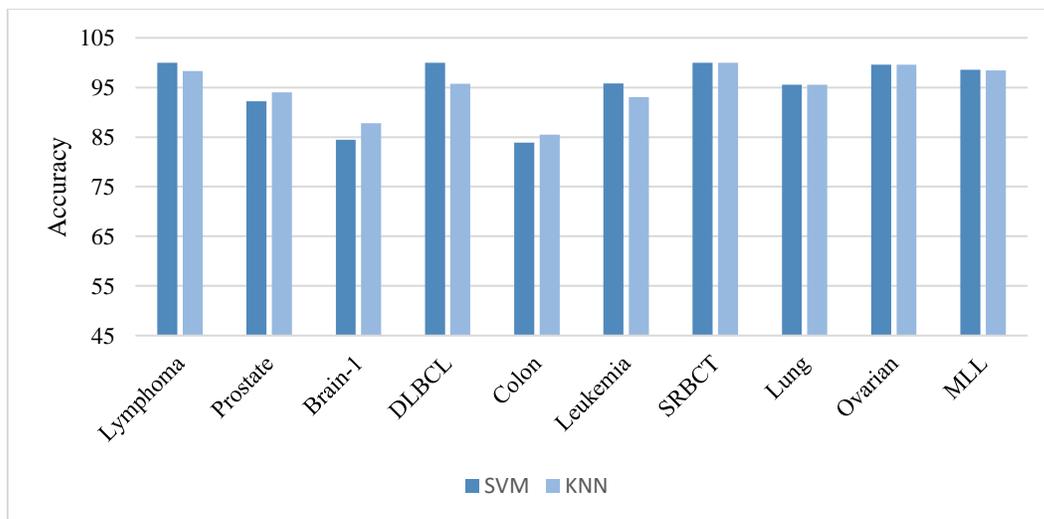

**Fig. 8** Comparison between the performance of two different classifiers on the obtained top 50 genes of each dataset by the MRMR filter method

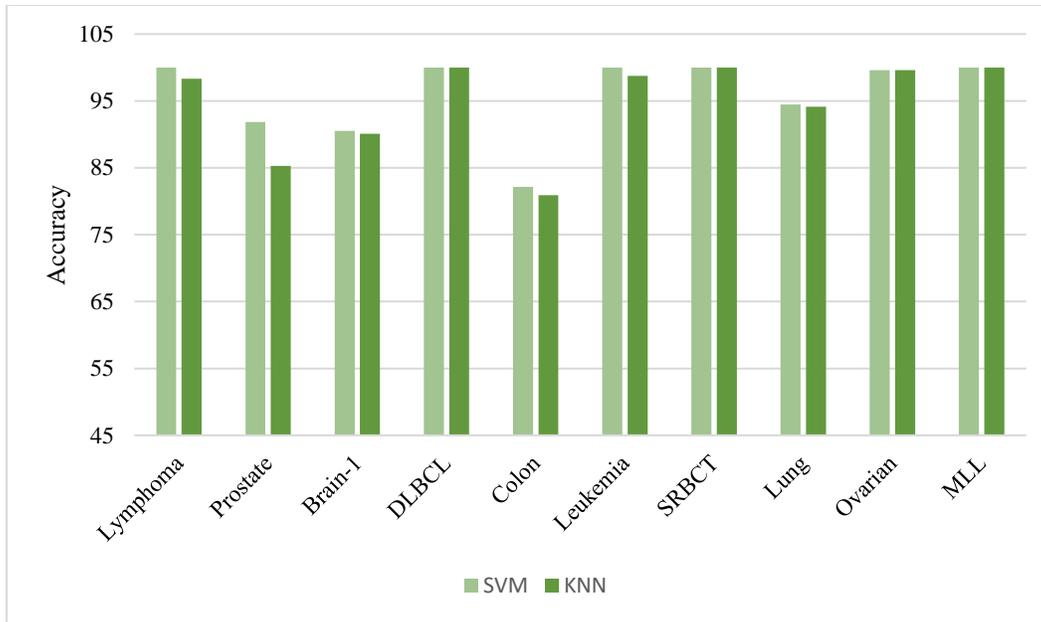

**Fig. 9** Comparison between the performance of two different classifiers on the obtained top 100 genes of each dataset by the MRMR filter method

### 5.4.2. Assessment of the performance of BHOA using S-Shaped and V-Shaped TFs

The effectiveness of binary HOA (BHOA) on feature selection is investigated using eight different transfer functions (S-Shaped and V-Shaped, as mentioned in Table 1) and SVM classifier with the linear kernel as an evaluator. Table 7 presents the results of BHOA based on S-Shaped and V-Shaped TFs in terms of best, mean, worst, and standard deviation of classification accuracy. Also, Table 8 reports the minimum number of selected genes using each transfer function. As shown in Tables 7 and 8, S-Shaped groups achieved the same results in 5 cases, and S1 showed good performance in 3 out of the rest datasets. Moreover, V1 obtained the highest classification accuracy in 5 out of 10 datasets in the V-Shaped family. Overall, the best results of S-Shaped families were reported by S1, and the V1 TF achieved higher average accuracy and a lower number of genes than other V-Shaped TFs on all datasets.

By comparing the results of S1 and V1 in terms of average accuracy, it can be found that they obtained the same results for five datasets, and V1 achieved higher performance on 3 out of 5 remaining datasets. From Table 8, it is observed that S1 and V1 obtained nearly the same results for selected genes. However, the results of S1 are better than V1 on 8 out of 10 datasets. Since the primary purpose of gene selection is to decrease the number of selected genes while maximizing classification accuracy, it can be concluded that BHOA using S1 performed well because they achieved the highest classification accuracy while using fewer genes. Additionally, comparing the standard deviations between S1 and V1 demonstrates S1 transfer function provided smoother results with enhanced stability.

**Table 7.** A comparison between the S-Shaped and V-Shaped transfer functions regarding the best, the mean, the worst, and STD of classification accuracy (%).

| Dataset | Performance | S-Shaped TF | | | | V-Shaped TF | | | |
|---------|-------------|-------|-------|-------|-------|-------|-------|-------|-------|
|         |             | S1 | S2 | S3 | S4 | V1 | V2 | V3 | V4 |
| Lymphoma | Best | 100 | 100 | 100 | 100 | 100 | 100 | 100 | 100 |
|          | Mean | 100 | 100 | 100 | 100 | 100 | 100 | 100 | 100 |
|          | Worst | 100 | 100 | 100 | 100 | 100 | 100 | 100 | 100 |
|          | STD | 0 | 0 | 0 | 0 | 0 | 0 | 0 | 0 |
| Prostate | Best | 95.61 | 95.5 | 95.16 | 94.35 | 93.59 | 92.98 | 93.71 | 92.99 |
|          | Mean | 95.59 | 95.02 | 95.03 | 94.01 | 93.12 | 92.61 | 92.82 | 92.05 |
|          | Worst | 95.28 | 94.51 | 94.78 | 93.49 | 92.46 | 92.01 | 92.40 | 91.69 |
|          | STD | 1.12 | 1.03 | 0.76 | 1.67 | 1.72 | 0.54 | 1.36 | 0.89 |
| Brain-1 | Best | 93.77 | 93.87 | 94.77 | 94.55 | 94.77 | 93.77 | 94.66 | 93.77 |
|          | Mean | 92.92 | 93.03 | 93.65 | 93.39 | 94.44 | 93.26 | 93.86 | 93.37 |
|          | Worst | 92.55 | 92.31 | 92.44 | 91.55 | 93.64 | 92.66 | 93.44 | 92.44 |
|          | STD | 0.48 | 0.73 | 0.82 | 1.31 | 0.45 | 0.55 | 0.47 | 0.53 |
| DLBCL | Best | 100 | 100 | 100 | 100 | 100 | 100 | 100 | 100 |
|       | Mean | 100 | 100 | 100 | 100 | 100 | 100 | 100 | 100 |
|       | Worst | 100 | 100 | 100 | 100 | 100 | 100 | 100 | 100 |
|       | STD | 0 | 0 | 0 | 0 | 0 | 0 | 0 | 0 |
| Colon | Best | 98.33 | 96.91 | 96.89 | 97.33 | 96.91 | 97.14 | 97.51 | 98.33 |
|       | Mean | 96.99 | 96.67 | 96.47 | 96.75 | 96.85 | 96.82 | 97.12 | 97.28 |
|       | Worst | 95.47 | 96.47 | 95.23 | 95.47 | 96.61 | 96.62 | 96.93 | 96.9 |
|       | STD | 1.01 | 0.15 | 0.71 | 0.73 | 0.14 | 0.19 | 0.27 | 0.61 |
| Leukemia | Best | 100 | 100 | 100 | 100 | 100 | 100 | 100 | 100 |
|          | Mean | 100 | 99.26 | 99.10 | 99.02 | 99.27 | 99.26 | 99.17 | 99.02 |
|          | Worst | 100 | 98.75 | 98.74 | 98.64 | 98.75 | 98.45 | 98.36 | 98.12 |
|          | STD | 0 | 0.64 | 0.60 | 0.71 | 0.64 | 0.69 | 0.71 | 0.73 |
| SRBCT | Best | 100 | 100 | 100 | 100 | 100 | 100 | 100 | 100 |
|       | Mean | 100 | 100 | 100 | 100 | 100 | 100 | 100 | 100 |
|       | Worst | 100 | 100 | 100 | 100 | 100 | 100 | 100 | 100 |
|       | STD | 0 | 0 | 0 | 0 | 0 | 0 | 0 | 0 |
| Lung | Best | 97.54 | 98.57 | 97.57 | 98.07 | 97.59 | 97.59 | 98.07 | 97.52 |
|      | Mean | 97.08 | 98.15 | 97.46 | 97.45 | 97.47 | 97.35 | 97.56 | 97.35 |
|      | Worst | 96.63 | 97.57 | 97.11 | 97.04 | 97.11 | 97.07 | 97.04 | 97.12 |
|      | STD | 0.32 | 0.41 | 0.19 | 0.42 | 0.25 | 0.25 | 0.36 | 0.37 |
| Ovarian | Best | 100 | 100 | 100 | 100 | 100 | 100 | 100 | 100 |
|         | Mean | 100 | 100 | 100 | 100 | 100 | 100 | 100 | 100 |
|         | Worst | 100 | 100 | 100 | 100 | 100 | 100 | 100 | 100 |
|         | STD | 0 | 0 | 0 | 0 | 0 | 0 | 0 | 0 |
| MLL | Best | 100 | 100 | 100 | 100 | 100 | 100 | 100 | 100 |
|     | Mean | 100 | 100 | 100 | 100 | 100 | 100 | 100 | 100 |
|     | Worst | 100 | 100 | 100 | 100 | 100 | 100 | 100 | 100 |
|     | STD | 0 | 0 | 0 | 0 | 0 | 0 | 0 | 0 |

**Table 8.** A comparison between the S-Shaped and V-Shaped transfer functions regarding the average and STD of the selected genes.

| Dataset | Measure | S-Shaped TF | | | | V-Shaped TF | | | |
|---------|---------|------|------|------|------|-------|-------|-------|-------|
| | | S1 | S2 | S3 | S4 | V1 | V2 | V3 | V4 |
| Lymphoma | Avg. | 17 | 17 | 17.16 | 18.8 | 15.2 | 17.33 | 15.6 | 17.12 |
| | STD | 1.33 | 1.62 | 2.41 | 1.30 | 2.2 | 1.2 | 2.1 | 2.5 |
| Prostate | Avg. | 23.6 | 23.2 | 24.6 | 23.79 | 24.89 | 25.07 | 25.73 | 25.40 |
| | STD | 1.94 | 2.16 | 1.98 | 1.49 | 2.06 | 2.16 | 1.65 | 1.01 |
| Brain-1 | Avg. | 26 | 24.4 | 22.8 | 23.8 | 25.8 | 22.4 | 21.6 | 22.8 |
| | STD | 1.22 | 1.14 | 1.64 | 0.84 | 1.31 | 1.52 | 2.07 | 1.92 |
| DLBCL | Avg. | 11.5 | 17.33 | 19.57 | 19.14 | 18.51 | 17.42 | 17 | 18.43 |
| | STD | 1.35 | 1.86 | 0.97 | 1.06 | 1.1 | 1.12 | 2.7 | 1.9 |
| Colon | Avg. | 22.4 | 24.6 | 22.2 | 24.2 | 27.4 | 26.83 | 24.2 | 24.8 |
| | STD | 1.89 | 1.52 | 0.84 | 1.48 | 1.67 | 1.92 | 1.09 | 1.78 |
| Leukemia | Avg. | 15.7 | 21.7 | 21.28 | 22.38 | 18.16 | 20.44 | 20.46 | 21.43 |
| | STD | 1.56 | 2.08 | 2.14 | 2.56 | 1.9 | 0.9 | 0.9 | 0.8 |
| SRBCT | Avg. | 11.8 | 17.75 | 19.2 | 19.33 | 17.33 | 18.42 | 18.87 | 18.68 |
| | STD | 0.87 | 1.28 | 1.31 | 1.21 | 1.6 | 1.5 | 1.4 | 1.3 |
| Lung | Avg. | 23 | 25.2 | 26 | 25 | 27 | 23 | 23.8 | 24.6 |
| | STD | 1.41 | 0.83 | 1.41 | 1.22 | 1.58 | 2.91 | 1.92 | 1.36 |
| Ovarian | Avg. | 17.66 | 17.87 | 17.94 | 17.89 | 18.77 | 18.81 | 18.75 | 18.82 |
| | STD | 2.06 | 1.64 | 1.87 | 1.59 | 1.3 | 1.3 | 1.5 | 1.4 |
| MLL | Avg. | 19.11 | 19.44 | 20.6 | 19.12 | 18.71 | 18.8 | 18.84 | 19.4 |
| | STD | 1.05 | 1.66 | 0.89 | 1.9 | 1.9 | 1.8 | 1.7 | 1.82 |

### 5.4.3. Assessment of the impact of X-Shaped TF on BHOA

To investigate the impact of the novel X-Shaped TF on the performance of the BHOA, we should compare the obtained results by BHOA and eight TFs and results by BHOA and X-Shaped TF. The best, mean, worst, and standard deviation (STD) of classification accuracy are common criteria used to evaluate the effectiveness of the proposed approach, and an average of selected genes is measured and displayed in Table 9. Then SVM classifier with 10-fold cross-validation is employed to evaluate the selected genes. It can be seen from Table 9 that the BHOA using X-Shaped TF is able to obtain 100% average accuracy for six datasets which are Lymphoma, DLBCL, SRBCT, Leukemia, Ovarian, and MLL. Also, it has achieved 98.48%, 98.66%, 97.72%, and 96.45% classification accuracy for the Colon, Lung, Prostate, and Brain-1 datasets, respectively. Furthermore, the proposed approach showed superior performance in terms of selected genes. In other words, the proposed method has the capability to gain higher accuracy with a minimum number of selected genes. To demonstrate the strength of X-Shaped TF, Table 10 shows the best results of X-Shaped, S1, and V1 TFs. According to Table 10, it can be detected that BHOA using X-Shaped TF shows superiority in comparison to S1 and V1 in terms of classification accuracy and the number of selected genes. It can be concluded that X-Shaped TF outperforms traditional transfer functions. Table 11 summarizes the best subset of genes for each dataset. Figs. 10 and 11 depict the convergence behavior of the proposed method when different transfer functions are used.

**Table 9**. Experiment results of MRMR-BHOA in terms of accuracy, the number of selected genes, and running time.

| Dataset | Accuracy (%) | | | | Number of Genes | | Running Time (s) |
|---|---|---|---|---|---|---|---|
| | Best | Mean | Worst | STD | Mean | STD | |
| Lymphoma | 100 | 100 | 100 | 0 | 2 | 0 | 44.28 |
| Prostate | 98.45 | 97.72 | 97.12 | 0.45 | 5.4 | 1.26 | 112.1 |
| Brain-1 | 96.89 | 96.45 | 95.75 | 0.49 | 8.81 | 1.25 | 94.34 |
| DLBCL | 100 | 100 | 100 | 0 | 2.8 | 0.63 | 31.85 |
| Colon | 100 | 98.48 | 96.66 | 0.87 | 7.36 | 1.61 | 53.91 |
| Leukemia | 100 | 100 | 100 | 0 | 2.6 | 0.51 | 34.88 |
| SRBCT | 100 | 100 | 100 | 0 | 5.53 | 1.05 | 60.96 |
| Lung | 99.21 | 98.66 | 98.49 | 0.23 | 8.36 | 2.20 | 114.82 |
| Ovarian | 100 | 100 | 100 | 0 | 3 | 0 | 52.53 |
| MLL | 100 | 100 | 100 | 0 | 4.1 | 0.85 | 48.13 |

**Table 10.** A comparison between the proposed method with S-Shaped and V-Shaped transfer functions regarding the average accuracy (%) and the number of selected genes.

| Dataset | Performance | Transfer Function | | |
|---|---|---|---|---|
| | | S-Shaped | V-Shaped | X-Shaped |
| Lymphoma | Accuracy | $100 \pm 0$ | $100 \pm 0$ | $\mathbf{100 \pm 0}$ |
| | #Gene | $17 \pm 1.33$ | $15.2 \pm 2.2$ | $\mathbf{2 \pm 0}$ |
| Prostate | Accuracy | $95.59 \pm 1.12$ | $93.12 \pm 1.72$ | $\mathbf{97.72 \pm 0.45}$ |
| | #Gene | $23.6 \pm 1.94$ | $24.89 \pm 2.06$ | $\mathbf{5.4 \pm 1.26}$ |
| Brain-1 | Accuracy | $92.92 \pm 0.48$ | $93.26 \pm 0.55$ | $\mathbf{96.45 \pm 0.49}$ |
| | #Gene | $26 \pm 1.22$ | $25.8 \pm 1.31$ | $\mathbf{8.81 \pm 1.25}$ |
| DLBCL | Accuracy | $100 \pm 0$ | $100 \pm 0$ | $\mathbf{100 \pm 0}$ |
| | #Gene | $11.5 \pm 1.35$ | $18.51 \pm 1.1$ | $\mathbf{2.8 \pm 0.63}$ |
| Colon | Accuracy | $96.99 \pm 1.01$ | $96.82 \pm 0.14$ | $\mathbf{98.48 \pm 0.87}$ |
| | #Gene | $22.4 \pm 1.89$ | $27.4 \pm 1.67$ | $\mathbf{7.36 \pm 1.61}$ |
| Leukemia | Accuracy | $100 \pm 0$ | $99.17 \pm 0.64$ | $\mathbf{100 \pm 0}$ |
| | #Gene | $15.7 \pm 1.56$ | $18.16 \pm 1.9$ | $\mathbf{2.6 \pm 0.51}$ |
| SRBCT | Accuracy | $100 \pm 0$ | $100 \pm 0$ | $\mathbf{100 \pm 0}$ |
| | #Gene | $11.8 \pm 0.87$ | $17.33 \pm 1.6$ | $\mathbf{5.53 \pm 1.05}$ |
| Lung | Accuracy | $97.08 \pm 0.32$ | $97.47 \pm 0.25$ | $\mathbf{98.66 \pm 0.23}$ |
| | #Gene | $23 \pm 1.41$ | $27 \pm 1.58$ | $\mathbf{8.36 \pm 2.20}$ |
| Ovarian | Accuracy | $100 \pm 0$ | $100 \pm 0$ | $\mathbf{100 \pm 0}$ |
| | #Gene | $17.66 \pm 2.06$ | $18.77 \pm 1.3$ | $\mathbf{3 \pm 0}$ |
| MLL | Accuracy | $100 \pm 0$ | $100 \pm 0$ | $\mathbf{100 \pm 0}$ |
| | #Gene | $19.11 \pm 1.05$ | $18.71 \pm 1.9$ | $\mathbf{4.1 \pm 0.85}$ |

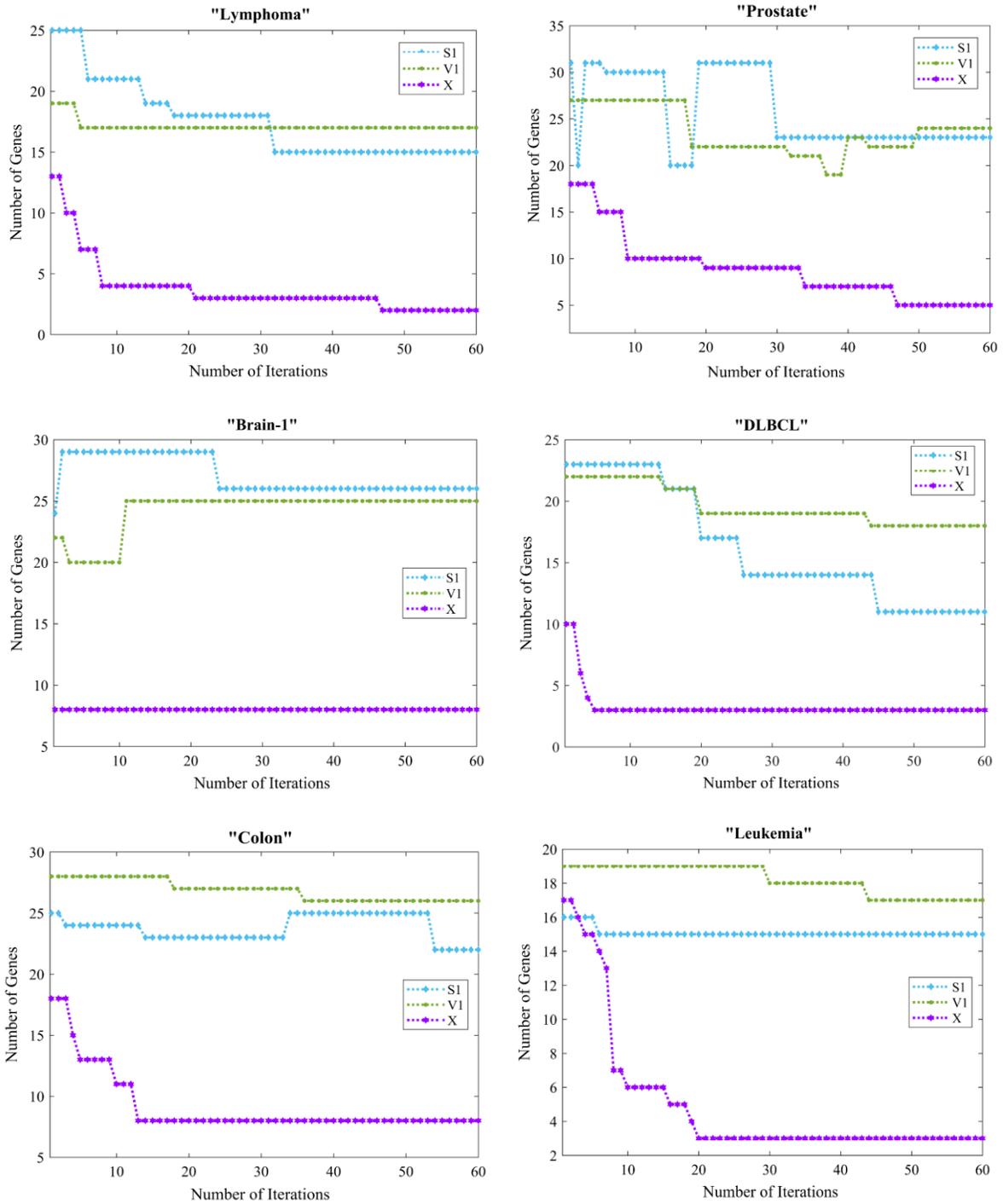

**Fig. 10** The convergence behavior of the proposed method using X-Shaped, S-Shaped, and V-Shaped TFs

**Table 11.** The obtained best subset of genes by the proposed approach for each dataset.

| Dataset | #Genes | Gene index | Gene name |
|---------|--------|-----------|-----------|
| Lymphoma | 2 | 8, 47 | V2750, V3754 |
| Prostate | 6 | 1, 3, 10, 11, 24, 45 | - |
| Brain-1 | 6 | 1, 10, 15, 37, 41, 50 | V2532, V5066, V3586, V2331, V5199, V1955 |
| DLBCL | 3 | 6, 12, 47 | V1281, V1291, V3020 |
| Colon | 8 | 2, 5, 12, 21, 24, 35, 42, 45 | - |
| Leukemia | 3 | 1, 19, 27 | V1834, V1685, V1630 |
| SRBCT | 5 | 10, 12, 14, 17, 18 | V1536, V255, V2050, V335, V417 |
| Lung | 7 | 8, 13, 21, 22, 29, 46, 47 | V8457, V8531, V8125, V4525, V9164, V12375, V5486 |
| Ovarian | 3 | 1, 17, 50 | V1680, V182, V2241 |
| MLL | 4 | 3, 9, 13, 17 | - |

## 5.5. Comparative results analysis between the proposed method and other meta-heuristic algorithms

This section provides a comparison between the BHOA algorithm and several popular meta-heuristic algorithms, including Gray Wolf Optimization (GWO), Whale Optimization Algorithm (WOA), Ant Colony Optimization (ACO), Particle Swarm Optimization (PSO), Genetic Algorithm (GA), and Firefly algorithm, in order to demonstrate the efficacy of the proposed approach using X-Shaped transfer function. All experiments were carried out 20 times independently to acquire confident results. Table 3 displays the parameters that were used to run these algorithms. Furthermore, the obtained results are outlined in Table 12 with regards to the average classification accuracy, standard deviation, and the number of selected genes. From Table 12, it can be observed that the BHOA and WOA have achieved 100% accuracy on 6 out of 10 datasets. However, BHOA outperforms WOA for all datasets regarding the minimum number of selected genes. Although BHOA provided similar results to PSO, GA, GWO, and WOA in terms of accuracy in the Ovarian dataset, it was able to identify fewer optimal genes when higher accuracy was taken into account. On the SRBCT dataset, the same competition can be seen. BHOA obtained a minimal number of genes compared to PSO, GWO, and WOA, while they obtained 100% accuracy. For the rest of the datasets, BHOA performs better than others regarding average accuracy and selected genes. Moreover, the standard deviation is low for both accuracy and selected genes across all datasets, and it is smoother than other approaches. In summary, the proposed method is capable of identifying the minimal optimal genes with greater accuracy. Based on the results of Table 12, it is obvious that the proposed approach outperforms the other meta-heuristics used in this study. Figs. 12 and 13 compare the proposed method and other meta-heuristic algorithms in terms of classification accuracy and the number of selected genes.

**Table 12.** Comparison between the proposed method and other meta-heuristics in terms of accuracy (%) and the number of selected genes.

| Dataset | Criteria | Algorithm | | | | | | |
|---|---|---|---|---|---|---|---|---|
| | | PSO | GA | Firefly | GWO | ACO | WOA | **BHOA** |
| Lymphoma | Acc. | 100 ± 0 | 98.18 ± 1.61 | 98.48 ± 0.96 | 98.49 ± 0.96 | 97.72 ± 2.83 | 100 ± 0 | **100 ± 0** |
| | #Gene | 8.3 ± 0.36 | 9.1 ± 1.31 | 13.5 ± 1.26 | 9.16 ± 1.94 | 4.83 ± 1.32 | 5.42 ± 0.38 | **2 ± 0** |
| | rank | (3) | (6) | (5) | (4) | (7) | (2) | (1) |
| Prostate | Acc. | 95.70 ± 1.7 | 97.43 ± 1.38 | 91.5 ± 0.71 | 96.58 ± 1.25 | 97.35 ± 1.92 | 94.78 ± 0.49 | **97.72 ± 0.45** |
| | #Gene | 14.3 ± 1.7 | 9.2 ± 1.06 | 26 ± 0.87 | 19.2 ± 1.38 | 18.42 ± 1.59 | 22.77 ± 1.68 | **5.4 ± 1.26** |
| | rank | (5) | (2) | (7) | (4) | (3) | (6) | (1) |
| Brain-1 | Acc. | 94.68 ± 1.49 | 88.11 ± 0.62 | 91.83 ± 0.62 | 86.47 ± 1.63 | 89.25 ± 3.56 | 92.75 ± 1.94 | **96.45 ± 0.49** |
| | #Gene | 15.36 ± 1.32 | 18.62 ± 1.83 | 21 ± 0.89 | 15.33 ± 1.75 | 16.33 ± 2.42 | 14.75 ± 1.23 | **8.81 ± 1.25** |
| | rank | (2) | (6) | (4) | (7) | (5) | (3) | (1) |
| DLBCL | Acc. | 100 ± 0 | 98.46 ± 1.13 | 99.08 ± 1.05 | 99.29 ± 1.09 | 97.23 ± 1.69 | 100 ± 0 | **100 ± 0** |
| | #Gene | 6.57 ± 1.27 | 18.3 ± 1.02 | 13.33 ± 1.90 | 11.5 ± 2.34 | 5.83 ± 0.75 | 9.6 ± 0.24 | **2.8 ± 0.63** |
| | rank | (2) | (6) | (5) | (4) | (7) | (3) | (1) |
| Colon | Acc. | 90.89 ± 1.58 | 86.79 ± 2.06 | 88.36 ± 1.26 | 88.49 ± 1.98 | 86.82 ± 1.58 | 91.67 ± 1.31 | **98.48 ± 0.87** |
| | #Gene | 12.43 ± 1.82 | 15.14 ± 1.89 | 13.26 ± 1.84 | 12.63 ± 1.79 | 9.33 ± 1.51 | 10.65 ± 1.26 | **7.36 ± 1.61** |
| | rank | (3) | (7) | (5) | (4) | (6) | (2) | (1) |
| Leukemia | Acc. | 99.79 ± 0.43 | 97.16 ± 1.79 | 97.42 ± 0.95 | 99.89 ± 0.54 | 97.68 ± 1.13 | 100 ± 0 | **100 ± 0** |
| | #Gene | 12.4 ± 1.34 | 15.5 ± 0.64 | 8.5 ± 1.76 | 8.45 ± 1.75 | 4.83 ± 1.32 | 8.33 ± 0.43 | **2.6 ± 0.51** |
| | rank | (4) | (7) | (6) | (3) | (5) | (2) | (1) |
| SRBCT | Acc. | 100 ± 0 | 97.30 ± 1.68 | 99.36 ± 0.70 | 100 ± 0 | 98.07 ± 2.18 | 100 ± 0 | **100 ± 0** |
| | #Gene | 13.2 ± 1.04 | 15.6 ± 1.93 | 16.33 ± 1.21 | 14.33 ± 0.51 | 8.8 ± 1.31 | 17 ± 1.01 | **5.53 ± 1.05** |
| | rank | (2) | (7) | (5) | (3) | (6) | (4) | (1) |
| Lung | Acc. | 89.78 ± 1.94 | 94.26 ± 1.04 | 94.91 ± 0.82 | 94.19 ± 0.62 | 93.95 ± 1.63 | 96.78 ± 1.45 | **98.66 ± 0.23** |
| | #Gene | 21.4 ± 1.37 | 24 ± 1.01 | 18 ± 1.09 | 16.75 ± 0.95 | 16.33 ± 3.07 | 18.85 ± 0.89 | **8.36 ± 2.20** |
| | rank | (7) | (4) | (3) | (5) | (6) | (2) | (1) |
| Ovarian | Acc. | 100 ± 0 | 100 ± 0 | 99.88 ± 0.17 | 100 ± 0 | 99.86 ± 0.18 | 100 ± 0 | **100 ± 0** |
| | #Gene | 5.6 ± 0.78 | 18.35 ± 0.61 | 11.5 ± 1.52 | 7.5 ± 0.54 | 3.6 ± 0.55 | 8.12 ± 0.68 | **3 ± 0** |
| | Rank | (2) | (5) | (6) | (3) | (7) | (4) | (1) |
| MLL | Acc. | 98.64 ± 1.62 | 98.36 ± 0.95 | 99.52 ± 0.76 | 98.84 ± 1.62 | 99.10 ± 1.12 | 100 ± 0 | **100 ± 0** |
| | #Gene | 8.76 ± 1.04 | 10.84 ± 1.72 | 12.83 ± 2.23 | 13.66 ± 1.63 | 7.66 ± 1.03 | 10.44 ± 1.12 | **4.1 ± 0.85** |
| | rank | (6) | (7) | (3) | (5) | (4) | (2) | (1) |
| Average rank | | (3.6) | (5.7) | (4.9) | (4.2) | (5.6) | (3) | **(1)** |

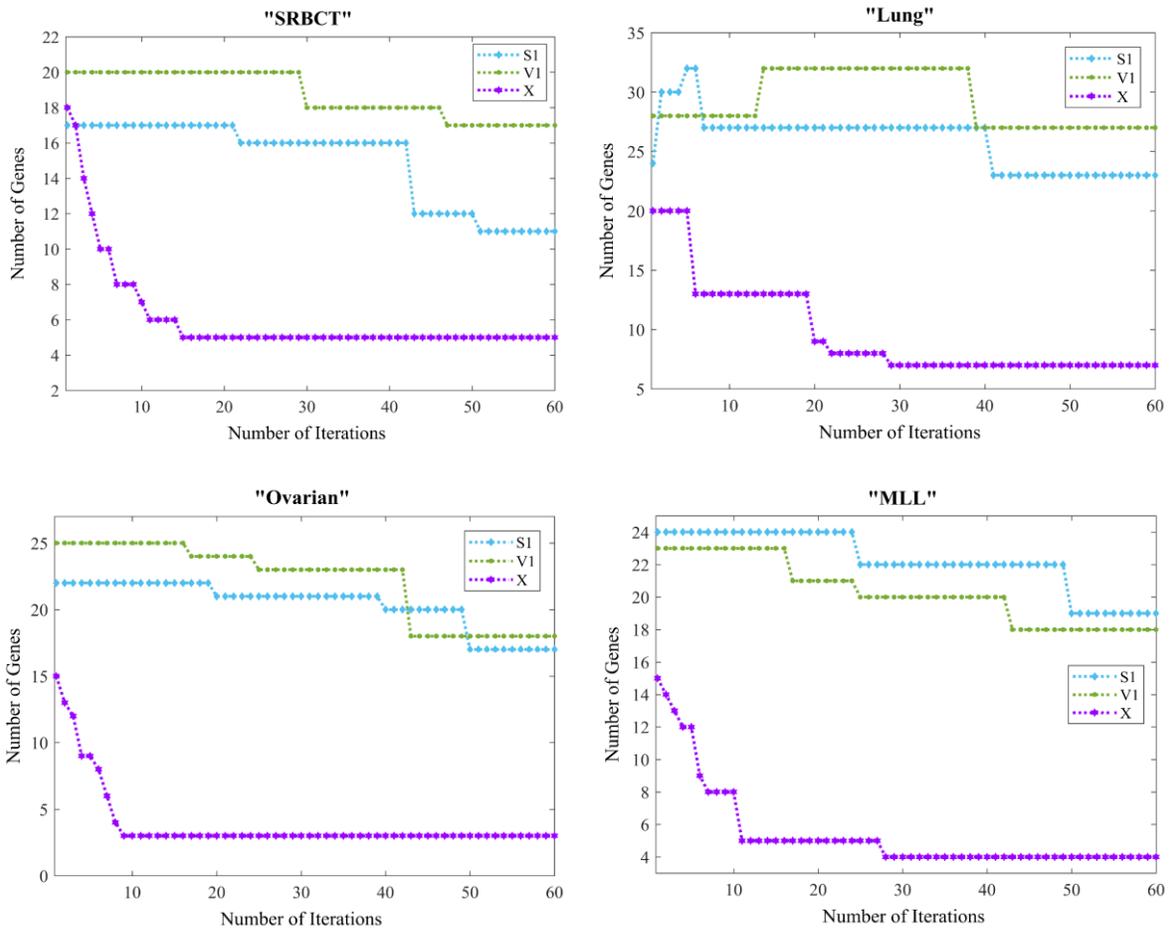

**Fig. 11.** The convergence behavior of the proposed method using X-Shaped, S-Shaped, and V-Shaped TF

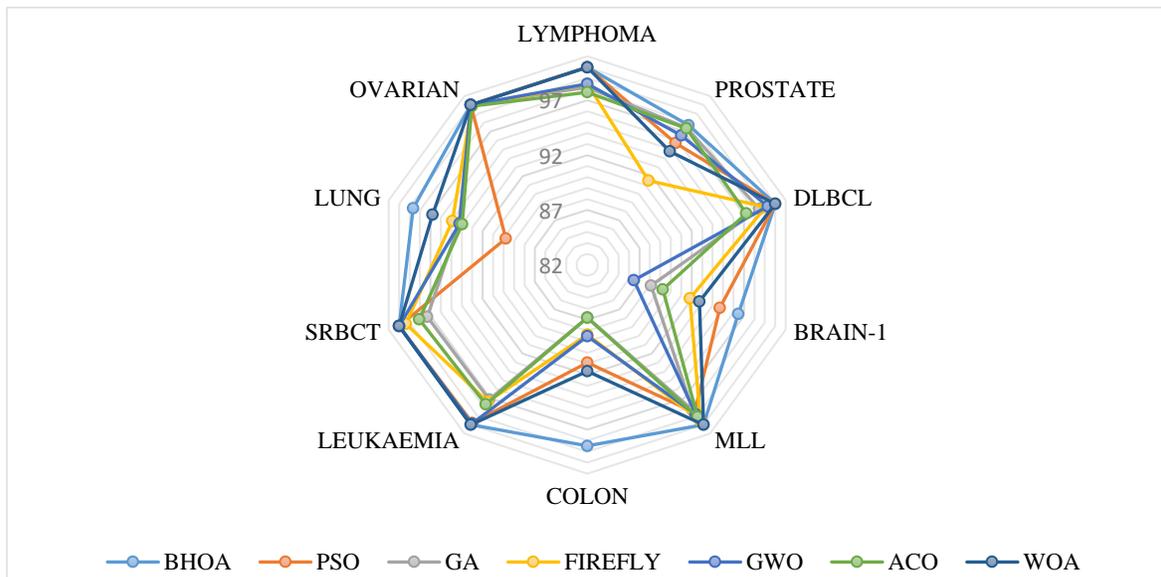

**Fig. 12** Comparison of proposed algorithm versus other meta-heuristic algorithms regarding the accuracy

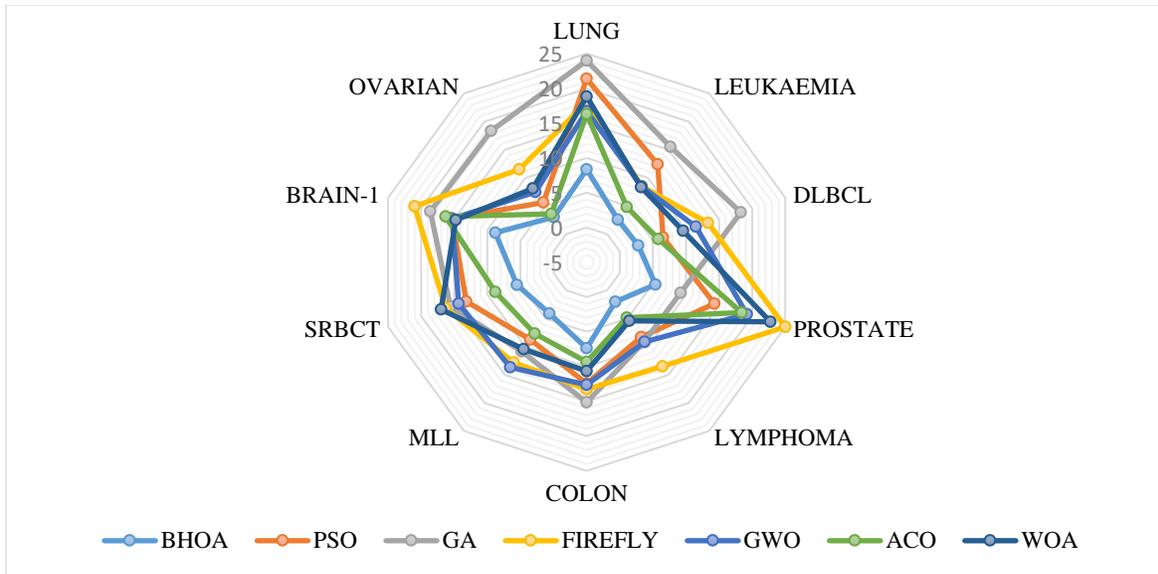

**Fig. 13** Comparison of proposed algorithm versus other meta-heuristic algorithms regarding selected genes

## 5.6. Evaluation of proposed method with past literature

Due to BHOA's superior performance compared to other meta-heuristic algorithms, experimental studies are expanded to compare MRMR-BHOA's performance with existing state-of-the-art algorithms. Table 13 summarizes the mean accuracy and the number of selected genes (in brackets) achieved by the proposed method and other methods in the literature. The bold type indicates the best results, and a "*" means no result was found. As can be seen from Table 13, MRMR-BHOA performed better than all other comparative methods for ten datasets concerning classification accuracy and the minimum number of selected genes.

To elaborate, MRMR-BHOA obtained 100% average accuracy with only two genes in Lymphoma dataset, while three approaches achieved 100% accuracy with more genes. For the Prostate dataset, BFO [62] and DRF0-CGS [63] as well as the proposed approach achieved approximately 97.5% accuracy. However, the proposed method identified only 5.4 genes, and as a result, it outperformed other techniques. In the Brain-1 dataset, DEFS [64] achieved 96.30% accuracy with 30 genes, while the proposed method obtained 96.45% with fewer genes (8.81) than DEFS. The proposed algorithm selected only 2.8 genes with 100% accuracy on the DLBCL dataset. For the colon dataset, the proposed algorithm provided higher accuracy with a smaller number of genes (7.36). All techniques have achieved 100% accuracy in the Leukemia dataset. On the other hand, the proposed method selected only 2.6 genes. Although 3 out of 5 methods obtained 100% accuracy similar to MRMR-BHOA, the proposed method chose fewer genes (5.53) than others on the SRBCT dataset. Moreover, DRF0-CGS and MRMR-BHOA performed better than the other methods in the lung dataset, despite this, MRMR-BHOA identified few genes (8.36). In the two remaining datasets, Ovarian and MLL, three cases of five methods have attained 100% accuracy, but the MRMR-BHOA algorithm has selected the fewest genes. For the Ovarian and MLL datasets, the proposed method reached 100% accuracy with only 3 and 4 generations, respectively. Overall, the findings indicate that the proposed methodology can produce fewer gene sets with more accurate classification than other approaches on the whole dataset.

**Table 13**. Comparison between the proposed approach and other methods in terms of accuracy (%) and the number of selected genes.

| Dataset | Method | Acc (#Gene) | Ref. | Dataset | Method | Acc (#Gene) | Ref. |
|---------|--------|-------------|------|---------|--------|-------------|------|
| Lymphoma | RMRMR-HBA | 100 (8.13) | [65] | Leukemia | RMRMR-HBA | 100 (4.07) | [65] |
| | BPSO-CGA | 100 (196) | [66] | | BPSO-CGA | 100 (300) | [66] |
| | BPSO-CGA | 82.14 (10.3) | [67] | | BPSO-CGA | 100 (4) | [68] |
| | BIRSW | 94 (30) | [68] | | DBH | 100 (16.1) | [70] |
| | IT-bBOA | 100 (24) | [69] | | TOPSIS-Jaya | 100 (4.3) | [69] |
| | CFC-iBPSO | **100 (2)** | | | CFC-iBPSO | **100 (2.6)** | |
| | **Proposed** | | | | **Proposed** | | |
| Prostate | IT-bBOA | 96 (30) | [68] | SRBCT | RMRMR-HBA | 100 (9.13) | [65] |
| | IBPSO | 92.16 (*) | [71] | | BPSO-CGA | 100 (880) | [66] |
| | BFO | 97.42 (29) | [62] | | BPSO-CGA | 100 (34.1) | [69] |
| | DRF0-CGS | 97.06 (*) | [63] | | CFC-iBPSO | 99.57 (8.9) | [73] |
| | IWSSr | 94.3 (10) | [72] | | HSA-MB | 99.23 (60.7) | [74] |
| | **Proposed** | **97.72 (5.4)** | | | MBEGA | **100 (5.53)** | |
| | | | | | **Proposed** | | |
| Brain-1 | BFO | 90.37 (25) | [62] | Lung | BFO | 93.11 (39) | [62] |
| | BPSO_TS | 95.89 (*) | [74] | | DRF0-CGS | 98.66 (17) | [63] |
| | IBPSO | 94.44 (*) | [71] | | PSO dICA | 97.95 (25) | [76] |
| | DEFS | 96.30 (30) | [64] | | BDE-X Rankf | 98.0 (3) | [77] |
| | IG-IBPSO | 92.22 (115) | [75] | | DLFCC | 83.00 (*) | [78] |
| | **Proposed** | **96.45 (8.81)** | | | **Proposed** | **98.66 (8.36)** | |
| DLBCL | DBH | 100 (4.05) | [70] | Ovarian | RMRMR-HBA | 100 (3.07) | [65] |
| | BFO | 98.99 (8) | [62] | | BPSO-CGA | 100 (3.3) | [69] |
| | DRF0-CGS | 94.67 (11) | [63] | | CFC-iBPSO | 99.81 (5.73) | [73] |
| | IWSSr | 94.73 (30) | [72] | | HSA-MB | 100 (2.6) | [70] |
| | BDE-X Rankf | 92.9 (3) | [77] | | DBH | 99.52 (18.5) | [79] |
| | **Proposed** | **100 (2.8)** | | | TOPSIS-Jaya | **100 (3)** | |
| | | | | | **Proposed** | | |
| Colon | RMRMR-HBA | 97.85 (12.27) | [65] | MLL | RMRMR-HBA | 100 (8) | [65] |
| | BPSO-CGA | 96.7 (214) | [66] | | CFC-iBPSO | 100 (30.08) | [69] |
| | DBH | 97.02 (14.4) | [70] | | HSA-MB | 99.55 (6.6) | [73] |
| | IT-bBOA | 86 (30) | [68] | | DBH | 100 (5.25) | [70] |
| | CFC-iBPSO | 94.89 (4.2) | [69] | | TOPSIS-Jaya | 99.62 (12.9) | [79] |
| | **Proposed** | **98.48 (7.36)** | | | **Proposed** | **100 (4.1)** | |

## 5.7. Statistical results

In order to substantiate the existence of significant differences between the results from multiple approaches, we used the Friedman test with 5% significance which is a commonly used non-parametric statistical method [80]. As outlined in Table 14, it generally ranks all methods separately on each dataset, the best algorithm receiving a rank of 1 and the worst algorithm receiving the last place. Tests are performed on accuracy and selected genes from each dataset. Tables 14 and 15 present the average rank of algorithms across all datasets using Friedman's statistics in terms of the classification accuracy and the number of selected genes. According to the results, the proposed method ranks first in both accuracy and selected genes. Based on the reported results, Friedman's statistics are 30.89999 and 40.5, and the p-values calculated in this experiment are 0.0000026488 and 0.0000000363 for accuracy and selected genes, respectively. In this case, the null hypothesis is rejected, which indicates that there is a difference between the performances of the compared algorithms. Since we found a statistically notable outcome, we used a post-hoc test for pairwise comparisons to find the precise cause of our differences. The p-value results of the post-hoc test between the classification accuracy and the minimum selected genes acquired by the BHOA and those of the PSO, GA, Firefly, ACO, GWO, and WOA meta-heuristic algorithms are exhibited in Tables 16 and 17. A p-value less than 0.05 confirms that the BHOA outperforms its corresponding optimization algorithm considerably. As a result, the proposed approach is statistically superior to the majority of existing methods. A comparison between the ranking of existing algorithms and the proposed approach by the Friedman test is illustrated in Fig. 14.

**Table 14.** Average ranking of accuracy values between 7 algorithms on ten biomedical datasets by Friedman test.

| Algorithm | Ranking |
| --- | --- |
| PSO | 2.9 |
| GA | 4.4 |
| Firefly | 3.9 |
| GWO | 3.2 |
| ACO | 4.3 |
| WOA | 1.9 |
| **Proposed** | **1** |

**Table 15.** Average ranking of selected genes between 7 algorithms on ten biomedical datasets by Friedman test.

| Algorithm | Ranking |
| --- | --- |
| PSO | 3.9 |
| GA | 5.8 |
| Firefly | 6 |
| GWO | 4.6 |
| ACO | 2.5 |
| WOA | 4.2 |
| **Proposed** | **1** |

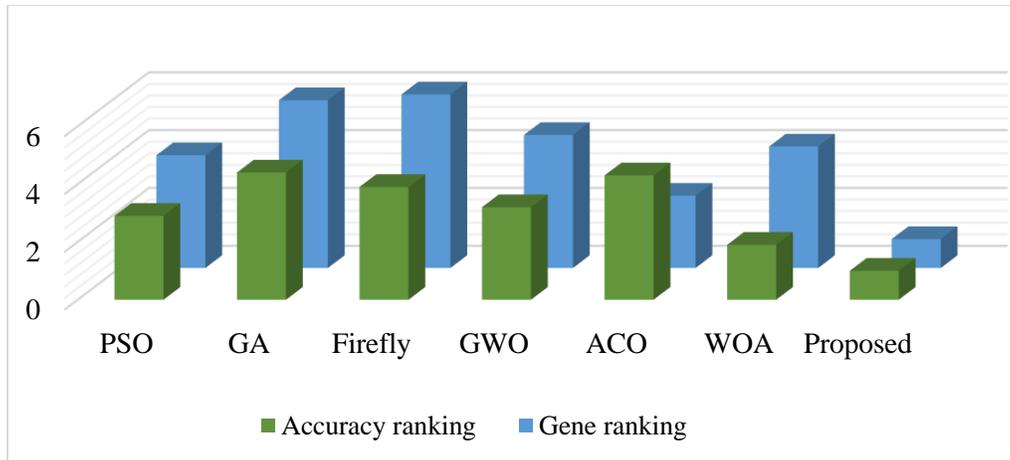

**Fig. 14** Comparison between the ranking of existing algorithms and the proposed approach by the Friedman test

**Table 16.** Post Hoc comparison regarding accuracy values.

| Comparison | Z-value | P-value | Result |
|---|---|---|---|
| Proposed vs PSO | -2.070197 | 0.038434 | $H_0$ is rejected |
| Proposed vs GA | -3.985129 | 0.000067 | $H_0$ is rejected |
| Proposed vs Firefly | -3.36407 | 0.000768 | $H_0$ is rejected |
| Proposed vs GWO | -2.587746 | 0.009661 | $H_0$ is rejected |
| Proposed vs ACO | -4.088638 | 0.000043 | $H_0$ is rejected |
| Proposed vs WOA | -0.931589 | 0.351549 | $H_0$ is not rejected |

**Table 17**. Post Hoc comparison regarding selected genes.

| Comparison | Z-value | P-value | Result |
|---|---|---|---|
| Proposed vs PSO | -3.001785 | 0.002684015 | $H_0$ is rejected |
| Proposed vs GA | -4.968472 | 0.000000675 | $H_0$ is rejected |
| Proposed vs Firefly | -5.175492 | 0.000000227 | $H_0$ is rejected |
| Proposed vs GWO | -3.726354 | 0.000194269 | $H_0$ is rejected |
| Proposed vs ACO | -1.552648 | 0.120507400 | $H_0$ is not rejected |
| Proposed vs WOA | -3.312315 | 0.000925274 | $H_0$ is rejected |

## 6. Conclusion

Over the past decades, biologists and life scientists have been fascinated by identifying significant genes related to cancer. However, finding meaningful genes in microarray datasets has always been difficult due to the abundance of genes and the scarcity of samples. In this study, gene selection as a binary problem has been tackled. The BHOA, a binary version of a new swarm intelligence algorithm inspired by horses' behavior, has been suggested to address the gene selection issue. In the proposed method, we use a hybrid of BHOA and MRMR, which benefits the wrapper-based and filter-based approaches. We first employed the MRMR method to eliminate redundant and irrelevant genes and recognize meaningful gene subsets. Afterward, continuous search space was mapped into binary values by using eight transfer functions that belong to the S-shaped and V-shaped families. On the other hand, an X-shaped TF was applied in MRMR-BHOA to improve the performance of BHOA in gene selection. Generally, X-Shaped TF employs the crossover operation to assist the optimization algorithm in identifying the appropriate region where the optimal solution may exist. When the efficacy of X-Shaped TF was tested against other TFs used, it was clearly superior in terms of obtaining the highest accuracy and finding the fewest number of genes. The proposed framework was evaluated across ten microarray datasets using an SVM classifier with 10-fold Cross-Validation. Furthermore, the capabilities of MRMR-BHOA were compared against six of the popular meta-heuristic algorithms. Consequently, the obtained results evidence that the proposed method has exhibited more notable performance than other approaches with regard to both classification accuracy and a minimum number of selected genes.